\documentclass[10pt,journal,compsoc]{IEEEtran}
\ifCLASSOPTIONcompsoc
  \usepackage[nocompress]{cite}
\else
  \usepackage{cite}
\fi
\usepackage{paralist}
\usepackage{graphicx}
\usepackage{comment}
\usepackage{amsmath,amssymb} 
\usepackage{color}
\usepackage{multirow}
\usepackage[export]{adjustbox}
\usepackage{array, booktabs}

\def\eg{\emph{e.g.}} 

\def\ie{\emph{i.e.}} 

\def\cf{\emph{cf. }}

\ifCLASSINFOpdf
\else
\fi
\begin{document}
%
\title{Displacement-Invariant Cost Computation\\ for Efficient Stereo Matching}
%
%
%
%

\author{Yiran~Zhong, Charles~Loop, Wonmin~Byeon, Stan~Birchfield, Yuchao~Dai,\\ Kaihao~Zhang, Alexey~Kamenev, Thomas~Breuel, Hongdong~Li, Jan~Kautz
\IEEEcompsocitemizethanks{\IEEEcompsocthanksitem Y.Zhong is with Australia National University. This work was done when he was an intern in NVIDIA, Redmond, WA 98052. C.Loop, W.Byeon, S.Birchfield, A.Kamenev, T.Breuel and J.Kautz are with NVIDIA.\protect\\
E-mail: yiran.zhong@anu.edu.au
\IEEEcompsocthanksitem Y.Dai is with Northwestern Polytechnical University. 
\IEEEcompsocthanksitem K.Zhang and H. Li are with Australia National University}
}

%
%

\markboth{Journal of \LaTeX\ Class Files,~Vol.~14, No.~8, August~2015}%
{Shell \MakeLowercase{\textit{et al.}}: Bare Demo of IEEEtran.cls for Computer Society Journals}
%



\IEEEtitleabstractindextext{%
\begin{abstract}
Although deep learning-based methods have dominated stereo matching leaderboards by yielding unprecedented disparity accuracy, their inference time is typically slow, on the order of seconds for a pair of 540p images.  The main reason is that the leading methods employ time-consuming 3D convolutions applied to a 4D feature volume. A common way to speed up the computation is to downsample the feature volume, but this loses high-frequency details. To overcome these challenges, we propose a \emph{displacement- invariant cost computation module} to compute the matching costs without needing a 4D feature volume. Rather, costs are computed by applying the same 2D convolution network on each disparity-shifted feature map pair independently. Unlike previous 2D convolution- based methods that simply perform context mapping between inputs and disparity maps, our proposed approach learns to match features between the two images. We also propose an entropy-based refinement strategy to refine the computed disparity map, which further improves speed by avoiding the need to compute a second disparity map on the right image. Extensive experiments on standard datasets (SceneFlow, KITTI, ETH3D, and Middlebury) demonstrate that our method achieves competitive accuracy with much less inference time. On typical image sizes, our method processes over 100 FPS on a desktop GPU, making our method suitable for time-critical applications such as autonomous driving. We also show that our approach generalizes well to unseen datasets, outperforming 4D-volumetric methods.
\end{abstract}

}

\maketitle

\IEEEdisplaynontitleabstractindextext

%
\IEEEpeerreviewmaketitle

\IEEEraisesectionheading{\section{Introduction}\label{sec:introduction}}

\IEEEPARstart{D}{eep} learning-based methods have achieved state-of-the-art on most of the standard stereo matching benchmarks (\ie, KITTI \cite{Menze2015CVPR}, ETH3D \cite{schoeps2017cvpr}, and Middlebury \cite{Scharstein14}). This success is achieved by aggregating information in a 4D feature volume (height $\times$ width $\times$ disparity levels $\times$ feature dimension), which is formed by concatenating each feature in one image with its corresponding feature in the other image, across all pixels and disparity levels. To process such a 4D volume, expensive 3D convolutions are utilized, thus making these methods significantly more time- and space-intensive than traditional approaches like semi-global matching (SGM) \cite{Hirschmuller2008} and its variants \cite{SGMEmbedded}.
For example, the traditional method known as embedded SGM~\cite{SGMEmbedded} achieves 100~FPS (frames per second) for a pair of 540p images, whereas most deep learning-based methods only manage about 2~FPS. 
Moreover, since this 4D feature volume grows with the cube of resolution, high-resolution depth estimation is prohibitively expensive.

\begin{figure}
\centering
\includegraphics[trim=5pt 12pt 0pt 0pt,width=1\linewidth]{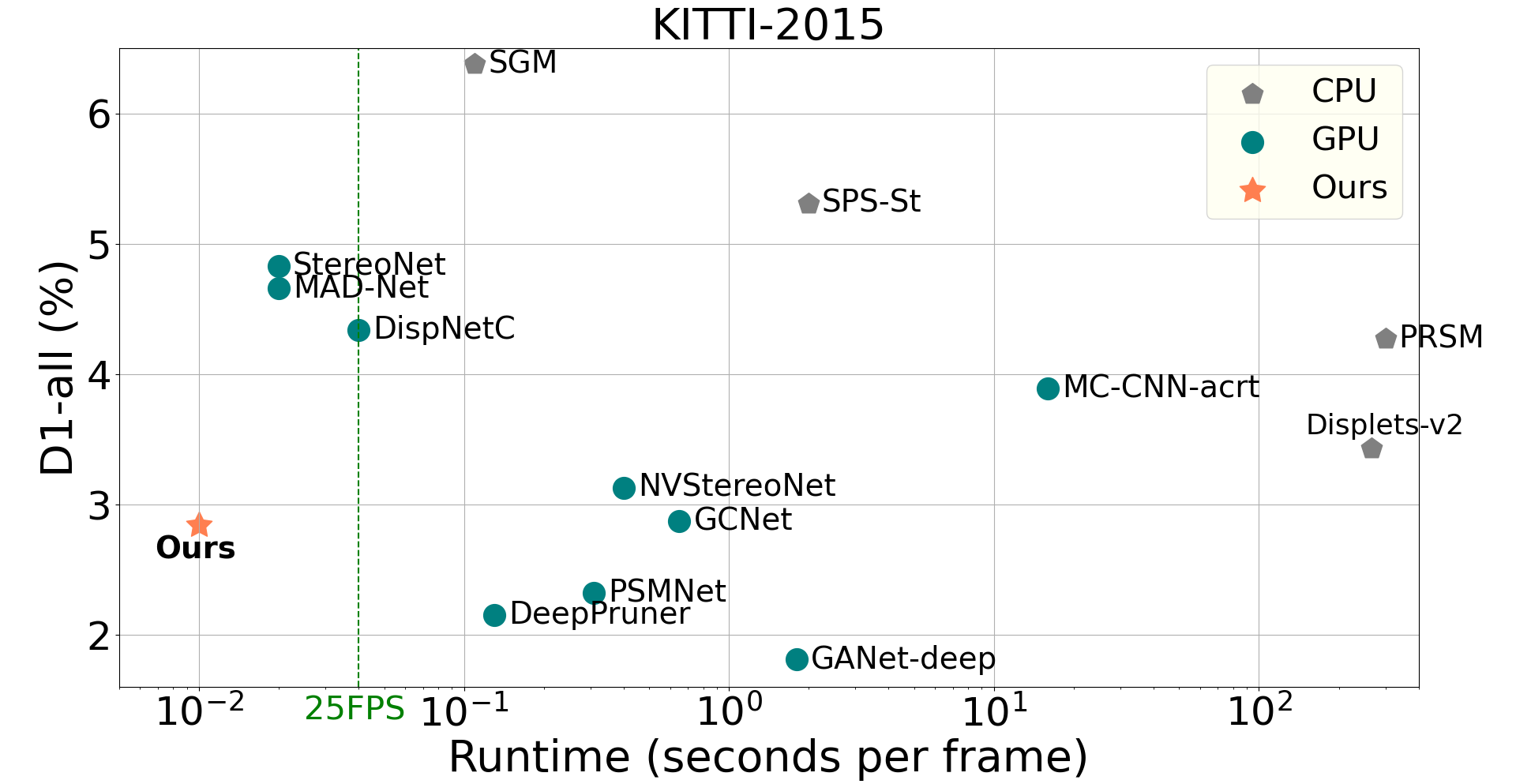}
\caption{{Our method achieves stereo matching accuracy comparable to state-of-the-art on the KITTI 2015 \emph{test} dataset, while operating at 100 FPS. (All GPU methods are timed on a NVIDIA RTX Titan GPU. Best viewed on screen.)}}
    \label{fig:accvsspeed}
\end{figure}

In this paper, we propose to overcome these limitations with a \emph{displacement-invariant cost computation module} that learns to match features of stereo images \emph{using only 2D convolutions}. Unlike previous 2D convolution-based methods \cite{MIFDB16,Liang2018Learning,pang2017cascade}, however, ours does not rely upon context matching between pixels and disparities; rather, due to its unique design, our network learns to match pixels between the two images.  
The key insight behind our approach is to compute the cost of each disparity shift independently using the same 2D convolution-based network. This vastly reduces the number of parameters and memory requirements for training and it also reduces time and memory costs during inference, as it does not need to explicitly store a 4D feature volume before cost computation. Also, we leverage entropy maps that computed from 3D cost volume as  confidence maps to guide the refinement of disparities. As a result (\cf, Fig.~\ref{fig:accvsspeed}), our proposed method is not only significantly faster than 3D convolution-based volumetric methods, \eg, GA-Net~\cite{Zhang2019GANet}, but it also achieves better cross-dataset generalization ability. The entire system is trained end-to-end.

Our paper contains the following contributions:
\begin{compactitem}
    \item[\small$\bullet$] A displacement-invariant cost computation module for stereo matching that uses 2D convolutions on disparity-shifted feature map pairs, which achieves significant speedup over standard volumetric approaches with much lower memory requirements.
    \item[\small$\bullet$] A new entropy based refinement scheme that bypasses the need for estimating disparity for the right view, which further reduces the processing time and memory consumption.  
    \item[\small$\bullet$] State-of-the-art performance on all three benchmarks compared with existing real-time methods, and better generalization performance than existing methods.
\end{compactitem}

\section{Related work}
Given a pair of rectified stereo images, stereo matching attempts to find a matching point for each pixel on the corresponding epipolar line.
Stereo matching has been extensively studied for decades in computer vision. Here we discuss a few popular and recent methods that are closely related to our approach. Interested readers are referred to recent survey papers such as \cite{Scharstein2002} and \cite{Geiger-Survey-2017}.

\subsection{Traditional Stereo Matching}
Traditional stereo matching methods can be roughly divided into two classes: local methods and global methods. The typical pipeline for a local method has four consecutive steps: 1) compute costs at a given disparity for each pixel; 2) sum up the costs over a window; 3) select the disparity that has the minimal cost; 4) perform a series of post-processing steps to refine the final results. 
Local methods \cite{Scharstein2002,Weber2009} have the advantage of speed. Since each cost within a window can be independently computed, these methods are highly parallelizable. The disadvantage of such methods is that they can only achieve sub-optimal results because they only consider local information. Global methods~\cite{meshstereo15,Taniai18} have been proposed to address this issue. They treat the disparity assignment task as a maximum flow / minimum cut problem and try to minimize a global cost function. However, such algorithms are typically too slow for real-time applications. Semi-global matching (SGM) \cite{Hirschmuller2008} is a compromise between these two extremes, which could achieve more accurate results than local methods without sacrificing speed significantly.

\subsection{Deep Stereo Matching}
Unlike traditional methods, deep stereo matching methods can learn to deal with difficult scenarios, such as repetitive textures or textureless regions, from ground truth data. A deep stereo method is often trained to benefit from one of the following aspects: 1) learning better features or metrics \cite{Zbontar2016};
2) learning better regularization terms \cite{Seki2017CVPR}; 
3) learning better refinement \cite{khamis2018stereonet}; 
4) learning better cost aggregation \cite{Zhang2019GANet}; 
5) learning direct disparity regression \cite{Mayer2016CVPR}. 
Based on the network structure, we divided these methods into two classes: 

\noindent\textbf{Direct Regression} methods often use an encoder-decoder to directly regress disparity maps from input images \cite{Mayer2016CVPR,Liang2018Learning,Tonioni_2019_CVPR,Yin_2019_CVPR}. Such methods learn in a brute force manner, discarding decades of acquired knowledge obtained by classical stereo matching research. When there are enough training data, and the train and test distributions are similar, such methods can work well. For example, iResNet~\cite{Liang2018Learning} won first place in the 2018 Robust Vision Challenge. Also, since such methods only employ 2D convolutions, they can easily achieve real-time or near real-time processing and have low GPU memory consumption. However, these methods lack the ability to generalize effectively, \eg, DispNet~\cite{Mayer2016CVPR} fails random dot stereo tests \cite{Zhong2018ECCV_rnn}.  

\noindent\textbf{Volumetric Methods} \cite{KendallMDHKBB17,Zhang2019GANet,SsSMnet2017,smolyanskiy2018nvstereo} build a 4D feature volume using features extracted from stereo pairs, which is then processed with multiple 3D convolutions. These methods leverage the concept of semi-global matching while replacing hand-crafted cost computation and cost aggregation steps with 3D convolutions. Since this type of network is forced to learn matching, volumetric methods easily pass random dot stereo tests \cite{Zhong2018ECCV_rnn}. However, they suffer from high processing times and high GPU memory consumption \cite{Zhang2019GANet}. 
To overcome these issues, some researchers build the feature volume at a lower image resolution to reduce memory footprint \cite{khamis2018stereonet,Yang_2019_CVPR}, or prune the feature volume by generating a confidence range for each pixel and aggregating costs within it \cite{Duggaliccv2019}. However, lowering the resolution of the feature volume makes it difficult to recover high-frequency information. To restore these lost details, volumetric networks typically use color-guided refinement layers as a final processing stage.

\subsection{Real-time Stereo Matching}
Several real-time deep stereo matching networks have been proposed recently. StereoNet \cite{khamis2018stereonet} achieves real-time performance at the cost of losing high frequency details due to a low resolution 4D cost volume, \ie, $1/8$ or $1/16$ of the input images. DispNet \cite{Mayer2016CVPR} and MADNet \cite{Tonioni_2019_CVPR} discard the general pipeline of stereo matching and design a context regression network that directly regresses disparity maps from the input images; these approaches suffer generalization problems as shown in \cite{Zhong2018ECCV_rnn}. 
To overcome the limitations of previous methods, our proposed approach uses a displacement-invariant cost computation module to achieve super real-time inference while, at the same time, preserving most high-frequency details by constructing a 3D cost volume on $1/3$ of the input resolution. Moreover, experiments in Section~\ref{sec:ab} show that our method achieves better generalization ability than 4D volumetric methods.

\section{Method}
\begin{figure*}[t]
    \centering
    \includegraphics[width=0.9\linewidth,clip,trim=0 1cm 0 1.5cm]{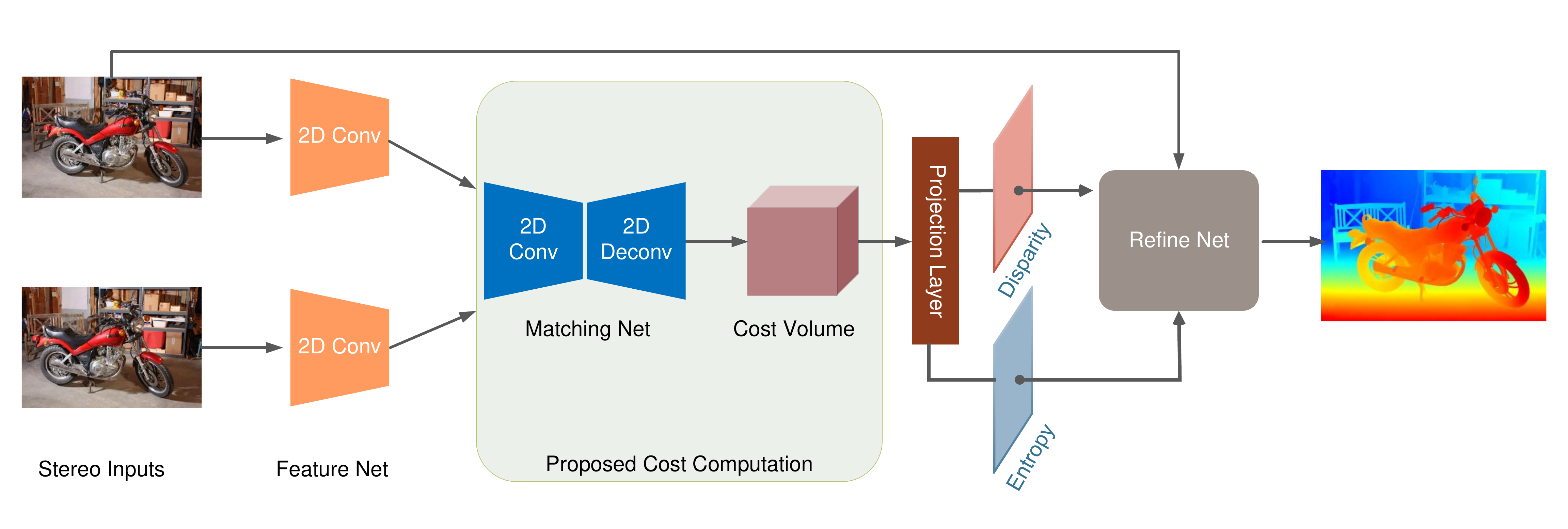}
    \vspace{-4mm}
    \caption{\textbf{Overall Architecture of our Stereo Network,} consisting of four components: Feature Net, Matching Net, Projection Layer and Refine Net. Given a pair of stereo images, the Feature Net produces feature maps which are processed by the Matching Net to generate a 3D cost volume (see Section~\ref{sec:dica}). The Projection Layer projects the volume to obtain a disparity map while also computing an entropy map. The Refine Net takes the left image and the output of the Projection Layer to generate the final disparity map.}
    \label{fig:structure}
\end{figure*}
In this section, we first describe the overall architecture and then provide a detailed analysis of our proposed \emph{displacement-invariant cost computation} module. 
Unlike existing volumetric methods, our method does not need to construct a 4D feature volume or use 3D convolutions for cost computation, which are the main barriers for applying volumetric methods to high-resolution images. By processing each disparity shift independently, our network only needs 2D convolutions to perform cost computation. We also propose a new refinement scheme using entropy, which bypasses the need to estimate disparity for the right view. 
\subsection{Network Architecture}
Fig.~\ref{fig:structure} illustrates the overall architecture of our framework, which consists of four major parts: (1) Feature Net, (2) Matching Net, (3) Projection Layer, and (4) Refine Net. 

\textbf{Feature Net.} A deeper feature net has a larger receptive field and can extract richer information. Therefore, researchers often use more than 15 layers for feature extraction \cite{KendallMDHKBB17,khamis2018stereonet,Zhang2019GANet}. 
In practice, we find that increasing the number of layers in the feature net does not help much for the final accuracy, \emph{e.g.}, MC-CNN \cite{Zbontar2016}. Therefore, in our feature net design, we use a shallow structure that only contains 8 convolutional layers. We first downsample the input images using a $3\times 3$ convolutions with a stride of 3. Three dilated convolutions are then applied to enlarge the receptive field. We also adopt a reduced spatial pyramid pooling (SPP) module \cite{SPP2014} to combine features from different scales to relieve the fixed-size constraint of a CNN. Our SPP module contains two average pooling layers: $64\times 64$ and $16\times 16$. Each of them follows a $1\times 1$ convolution and a bilinear upsampling layer. We concatenate feature maps that feed into our SPP module, then pass them to a $3\times 3$ convolution with output channel size of 96. The final feature map is generated by $1\times 1$ convolution without batch normalization and activation functions. Following previous work \cite{KendallMDHKBB17,khamis2018stereonet,Zhang2019GANet}, we use 32-channel feature maps.


\textbf{Matching Net.} Our Matching Net computes a cost map on each disparity level. We adopt a skip-connected U-Net \cite{Unet2015} with some modifications. In particular, we downsample each concatenated feature map four times using $3\times 3$ convolutions with a stride of 2. For each scale, we filter the feature maps with one $3\times 3$ convolution followed by a batch normalization layer and a ReLU activation layer. We set the feature size of each scale at 48, 64, 96 and 128 respectively. For upsampling layers, we use bi-linear upsampling and a $3\times 3$ convolution layers with a stride of 1 and reduce the feature dimension accordingly. A $3\times 3$ convolutional layer with feature size of 1, with no batch normalization nor activation applied to generate the final cost map. 

\setlength{\intextsep}{0pt}
\begin{figure}
 \centering
    \tabcolsep=0.05cm
    \begin{tabular}{c c c}
\small Left Image & \small Entropy Map \\
    \includegraphics[width=0.49\linewidth]{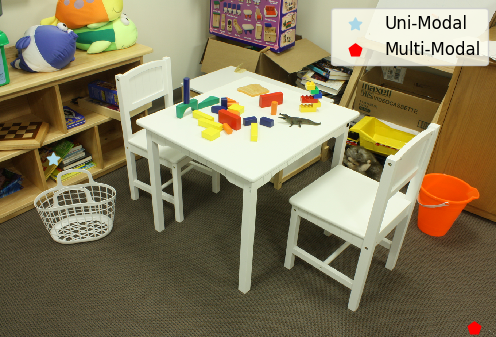} 
    &\includegraphics[width=0.49\linewidth]{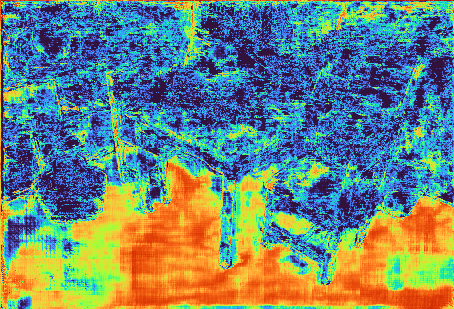} \\
    \end{tabular} \\
    \includegraphics[width=1\linewidth]{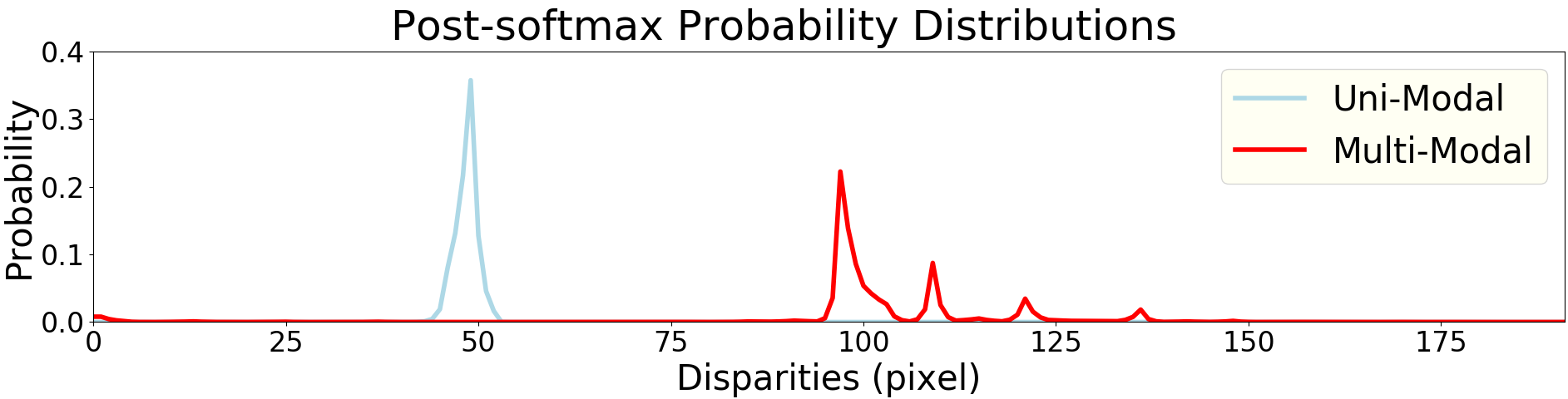} 
    \caption{\small \textbf{Entropy Map Analysis.} In the entropy map, red color corresponds to high entropy value. Best viewed in color.}
    \label{fig:entropy}
\end{figure}

\textbf{Projection layer.} After building a 3D cost volume, we use a projection layer to select the disparity with the lowest matching cost. Similar to previous volumetric methods \cite{KendallMDHKBB17,Zhong2018ECCV_rnn,Zhang2019GANet}, we use the soft-argmin operation to generate a disparity. The soft-argmin operation is defined as:
\begin{equation}
    \hat d := \sum_{d=0}^D[d\times \sigma(-c_d)],
\end{equation}
where $c_d$ is the matching cost at disparity $d$, $D$ is the preset maximum disparity level and $\sigma(\cdot)$ denotes the softmax operator. 

With little additional computation, the projection layer also generates an entropy map to estimate the confidence of each pixel. The entropy of each pixel is defined as:
\begin{equation}
    h = -\sum_{d=0}^D\sigma(-c_d)\log(\sigma(-c_d)).
\end{equation}
We apply a softmax operation on the cost to convert it to a probability distribution before computing the entropy. 
Fig.~\ref{fig:entropy} shows the resulting entropy map: pixels on textureless areas (\eg, red dot in the carpet) have high entropy, whereas the pixels on the texture-rich areas (\eg, blue dot in the shelf) have low entropy.
The bottom of the figure shows the post-softmax probability distribution curves of two selected pixels. A unimodal curve indicates low entropy (high confidence) for the pixel's estimated disparity, whereas a multimodal curve indicates high entropy (low confidence). 

\textbf{Refine Net.}
There are several choices in designing our refine net. StereoNet \cite{khamis2018stereonet} proposes a refine net that takes the raw disparity map and the left image as inputs. StereoDRNet \cite{Chabra_2019_CVPR} uses a similar refine net but takes the disparity map, the left image, image warping error map and disparity warping error map as inputs. The main drawback of such a design is that computing these error maps requires both left and right disparity maps, which costs extra time and memory. As shown in Fig.~\ref{fig:entropy}, the entropy map can provide similar information with less computation and memory. Therefore, we use a similar refine net architecture as StereoDRNet \cite{Chabra_2019_CVPR} but use the disparity map, left image and entropy map as inputs.

\subsection{Displacement-Invariant Cost Computation}
\label{sec:dica}
\begin{figure}[t]
    \centering
    \includegraphics[width=.9\linewidth]{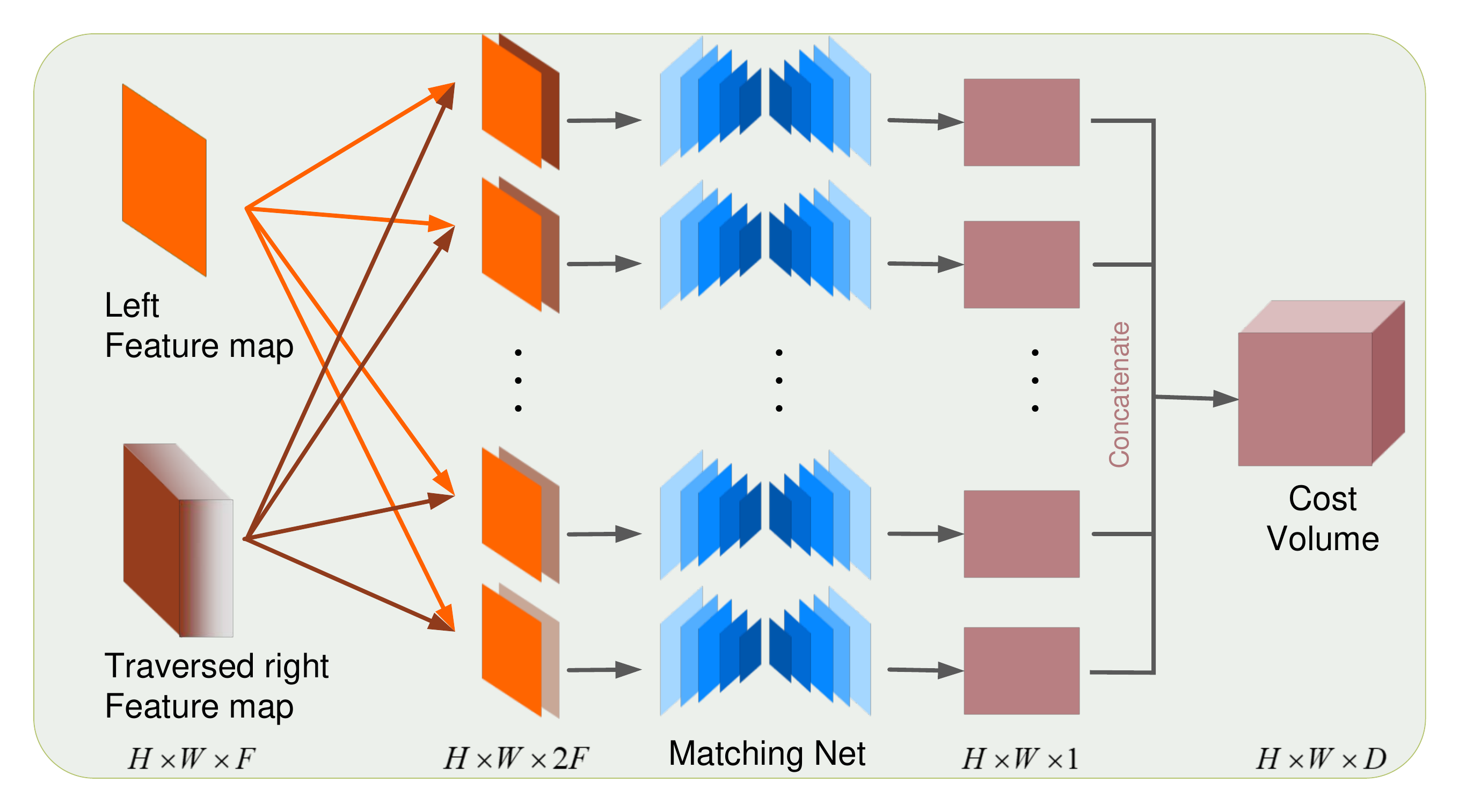}
    \vspace{-5mm}
    \caption{\textbf{Proposed method.} Our method runs share-weighted Matching Net on disparity shifted feature maps using 2D convolutions before 3D cost volume formation. In contrast, previous volumetric methods perform matching using 3D convolutions after constructing a 4D feature volume.}
    \label{fig:costval}
\end{figure}

Existing volumetric methods construct a 4D feature volume, which is processed with 3D convolutions to learn the matching cost, as shown in Fig.~\ref{fig:costval} (a). With such an approach, the matching cost for pixel $\mathbf{p}$ at disparity level $d$ for such approaches is calculated as:
\begin{equation}
    c_{3D}(\mathbf{p},d) = g_{3D}(\phi_{4D}(f(I^L(\mathbf{p})) \,\|\, f(I^R(\mathbf{p}-d))))
    \label{eq:3dmatch}, 
\end{equation}
%
where $f(\cdot)$ is a feature network to convert images to feature maps, $\phi_{4D}(\cdot \| \cdot)$ denotes the concatenation of disparity-shifted feature map pairs on every possible disparity shift, and $g_{3D}(\cdot)$ is a 3D convolution-based matching network that computes and aggregates the matching cost based on feature maps and neighboring disparity shifts. Therefore, the cost will be different if we shuffle the concatenated feature maps along the disparity dimension.

In contrast, our proposed Displacement-Invariant Cost Computation (DICC) only requires 2D convolutions. As shown in Fig.~\ref{fig:costval} (b), the exact same matching net is applied to each disparity-shifted feature map pair, thus ensuring that cost computation is only dependent on the current disparity shift. Unlike 3D convolution-based methods, our matching cost will be identical if we shuffle the concatenated feature maps along the disparity dimension. 



Specifically, let us consider a matching cost computation for pixel $\mathbf{p}$ at disparity level $d$. In our method, we compute the matching cost as follows:
\begin{equation}
    c_{2D}(\mathbf{p},d) = g_{2D}(f(I^L(\mathbf{p})), f(I^R(\mathbf{p}-d))),
    \label{eq:2dmatch} 
\end{equation}
where $f(\cdot)$ is a feature net, and $g_{2D}(\cdot)$ is a matching net that computes the matching cost at each disparity level independently. 
Similar to MC-CNN~\cite{Zbontar2016}, our proposed approach learns a cost from a deep network. However, MC-CNN computes the matching cost based on patches and then uses traditional semi-global cost aggregation steps to generate the disparity map. Our approach, on the other hand, is an end-to-end method that uses a network to perform cost computation and aggregation simultaneously. 


To better understand and compare the matching cost computation, we implement \emph{Baseline3D} using Equation~\eqref{eq:3dmatch} for cost computation, and we implement \emph{Ours} using Equation~\eqref{eq:2dmatch} to compute the matching cost. Baseline3D and Ours networks share the same feature net and projection layer, but Baseline3D constructs a 4D feature volume as in Fig.~\ref{fig:costval} (a) and replaces the 2D convolutions in our Matching Net with 3D convolutions.  A detailed comparison can be found in Section~\ref{sec:ab}.

\begin{table}[t]
\scriptsize
\centering
\caption{\small \textbf{Computational resource comparison between Our Matching Net and Baseline3D Matching Net}. Runtime is measured on $540 (H)\times 960 (W)$ resolution stereo pairs with $192 (D)$ disparity levels.}
\tabcolsep=0.05cm
\begin{tabular}{c c c c c}
\toprule
Methods & Runtime(s) & Params & Memory & Operations  \\ 
\hline
3D Matching Net & 0.150  &  3.84M  & $\frac{1}{3}H\times \frac{1}{3}W\times \frac{1}{3}D\times 2F$ & 733.8~Gflops\\
Our Matching Net & 0.001  &  1.16M  & $\frac{1}{3}H\times \frac{1}{3}W\times 2F$ & 5.4~Gflops\\
\bottomrule
\end{tabular}
\label{tab:2d3d}
\end{table}

In Table~\ref{tab:2d3d}, we compare the computational resources needed for Baseline3D and our proposed Matching Net. Note that in order to compute matching cost for each disparity level, we need to run our Matching Net $\frac{1}{3}D$ times. Since all disparity levels are processed independently, we are allowed to run our Matching Net either in parallel or sequentially. To construct a cost volume of size $\frac{1}{3}H\times \frac{1}{3}W\times \frac{1}{3}D$, the former one will need a memory of $\frac{1}{3}H\times \frac{1}{3}W\times \frac{1}{3}D\times 2F$ but can finish in $0.007s$ (including memory allocation time). The latter one just needs a memory of $\frac{1}{3}H\times \frac{1}{3}W\times 2F$ as all Matching Nets can share the same memory space and can finish in $0.07s$. We ignore the memory cost of convolution layers here as they may vary on different architectures. It is worth noting that the main obstacle for training a volumetric network on high resolution stereo pairs is that it requires the network to initialize a giant 4D feature volume in GPU memory before cost computation and the size of the feature volume increases as the cube of the inputs. However, since the 4D feature volume is not required for our method, we can handle high resolution inputs and large disparity ranges, \ie, our method can process a pair of $1500\times 1000$ images with 400 disparity levels in Middlebury benchmark without further downsampling the feature size to $1/64$ as in HSM \cite{Yang_2019_CVPR}.

Another advantage of our method is better generalization capability. Unlike volumetric methods that utilize 3D convolutions to regularize $H,W,D$ dimensions simultaneously, we force our method to learn matching costs only from $H,W$ dimensions. we argue that connections between $H\times W$ plane and $D$ dimension is a double-edged sword in computing the matching costs. While such connections do improve results on a single dataset, in practice we find that they lead to higher cross-dataset errors as verified in Section~\ref{sec:ab}.


\subsection{Loss Function}
Following GA-Net \cite{Zhang2019GANet}, we use the smooth $\ell_1$ loss as our training loss function, which is robust at disparity discontinuities and has low sensitivity to outliers or noise. Our network outputs two disparity maps: a coarse prediction $\mathbf{d}_{\mathrm{coarse}}$ from the soft-argmin operation and a refined one $\mathbf{d}_{\mathrm{refine}}$ from our refine net. We apply supervision on both of them.
Given the ground truth disparity $\mathbf{d}_{gt}$, the total loss function is defined as: 
\begin{equation}
\mathcal{L}_{\mathrm{total}} = \ell(\mathbf{d}_{\mathrm{coarse}} - \mathbf{d}_{\mathrm{gt}}) +  \lambda \ell(\mathbf{d}_{\mathrm{refine}} - \mathbf{d}_{\mathrm{gt}}),
\label{eq:total}
\end{equation}
where $\lambda = 1.25$, and
\begin{equation}
\ell(x) = \left\{
             \begin{array}{lc}
             0.5x^2, &  |x| < 1\\
             |x| - 0.5, & \mathrm{otherwise}.  
             \end{array}
\right.
\label{eq:l1}
\end{equation}

\section{Experimental Results}
We implemented our stereo matching network in PyTorch. We used the same training strategy and data argumentation as described in GA-Net \cite{Zhang2019GANet} for easy comparison. Our network was trained in an end-to-end manner with the Adam optimizer ($\beta_1=0.9$, $\beta_2=0.999$). We randomly cropped the input images to $240\times 576$ and used a batch size of 104 on 8 Nvidia Tesla V100 GPUs. We pre-trained our network from random initialization and a constant learning rate of 1e-3 on the SceneFlow dataset for 60 epochs. Each epoch took around 15 minutes and the total pre-training process took around 15 hours. For inference, our network can run 100 FPS on NVIDIA Titan RTX and occupies less than 2~GB memory for a pair of 384$\times$1280 stereo images. 
We add different disparity levels as a batch and run the Matching Net over it to achieve parallelism. 
We comprehensively studied the characteristics of our network in ablation studies and evaluated our network on leading stereo benchmarks.
 
\subsection{Datasets}
Our method was evaluated on the following datasets:

\textbf{SceneFlow dataset.}
SceneFlow \cite{Mayer2016CVPR} is a large synthetic dataset containing 35454 training and 4370 testing images with a typical image dimension of $540\times 960$. This dataset provides both left and right disparity maps, but we only use the left ones for training. Note that for some sequences, the maximum disparity level is larger than a pre-set limit (192 for this dataset), so we exclude these pixels in our training and evaluation. We use this dataset for pre-training and ablation studies.

\textbf{KITTI 2015 dataset.}
KITTI 2015 contains 200 real-world drive scene stereo pairs for training and 200 for testing. They have a fixed baseline but the focal lengths could vary. The typical resolution of KITTI images is $376\times 1240$. The semi-dense ground truth disparity maps are generated by Velodyne HDL64E LiDARs with manually inserted 3D CAD models for cars \cite{Menze2015CVPR}. From the training set, we randomly select 160 frames for training and 40 frames for validation. We use a maximum disparity level of 192 in this dataset. We report our results on the test set benchmark, whose ground truth is withheld. 

\textbf{ETH3D stereo dataset.}
ETH3D is a small dataset that contains 27 images for training and 20 for testing, for both indoor and outdoor scenes. It is a challenging dataset in that most stereo pairs have different lighting conditions. In other words, the Lambertian assumption for stereo matching may not hold in some areas. Moreover, unlike all the other datasets which have color inputs, ETH3D is grayscale. It requires the network to handle different channel statistics and extract features that are robust to lighting conditions. The maximum input resolution is $576\times 960$. Since the maximum disparity of this dataset is very small, we reduce the maximum disparity level to 48 for this dataset.

\textbf{Middlebury 2014 stereo dataset.}
Middlebury 2014 is another small dataset. It contains 15 images for training and 15 for testing. This dataset is challenging for deep stereo methods not only because of the small size of the training set, but also due to the high resolution imagery with many thin objects. The full resolution of Middlebury is up to $3000\times 2000$ with $800$ disparity levels. To fit in GPU memory, most deep stereo methods can only operate on quarter-resolution images. As a consequence, many details are missed at that resolution which leads to a reduced accuracy. We use half resolution and 432 disparity levels.  
 
\subsection{Evaluation on Benchmarks}
In this section, we provide results on three well-known stereo matching benchmarks: KITTI 2015, ETH3D, and Middlebury. We fine-tuned the SceneFlow pre-trained model on each benchmark. The algorithms are divided into two categories based on runtime: below 10 FPS and above 10 FPS.

\begin{figure}[t]
    \centering
    \tabcolsep=0.01cm
    \begin{tabular}{c c c}
    \scriptsize RGB & \scriptsize DeepPruner & \scriptsize Ours\\
    \includegraphics[width=0.325\linewidth]{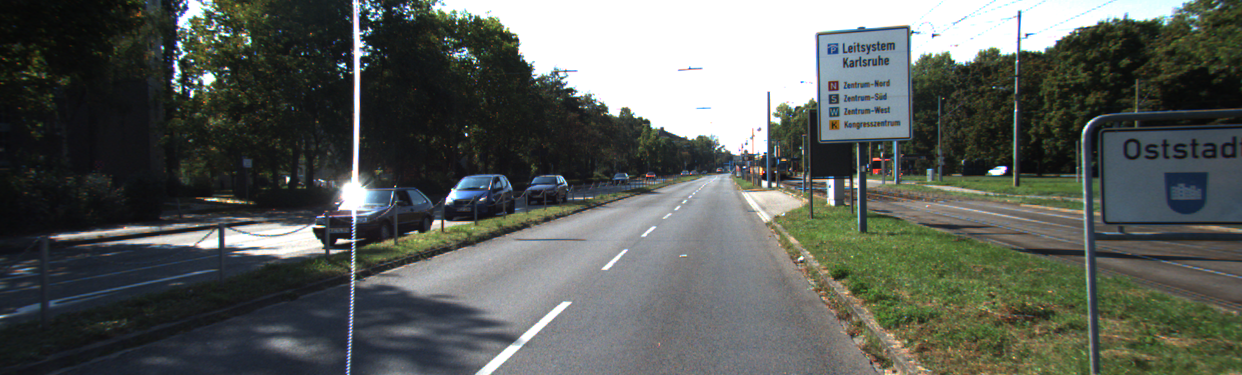}
    &\includegraphics[width=0.325\linewidth]{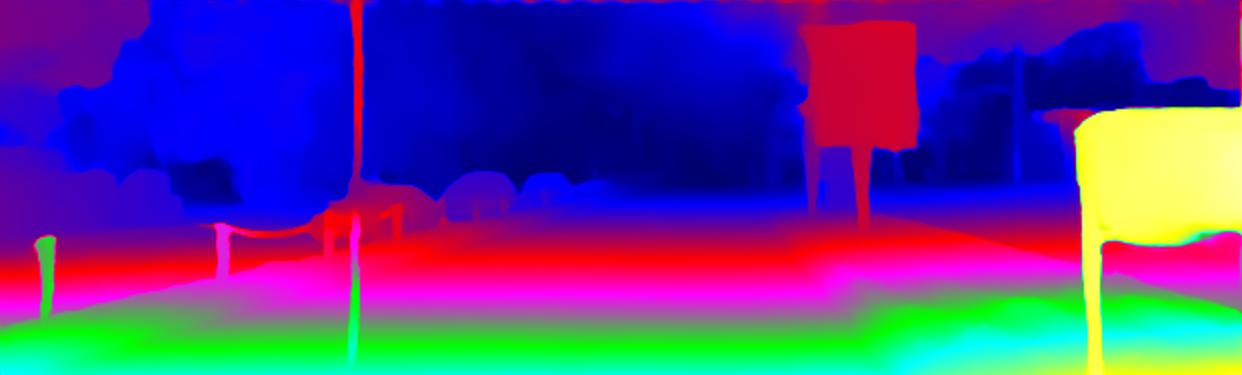}
    &\includegraphics[width=0.325\linewidth]{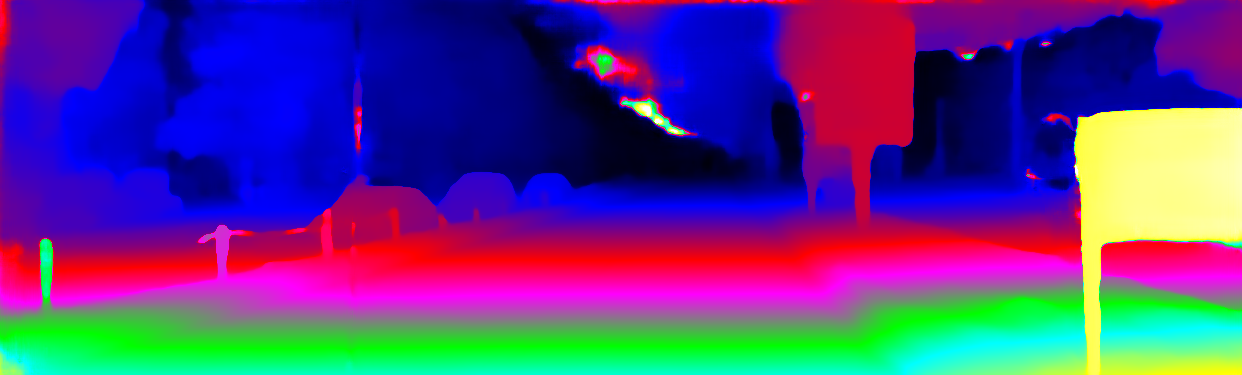}\\
    \includegraphics[width=0.325\linewidth]{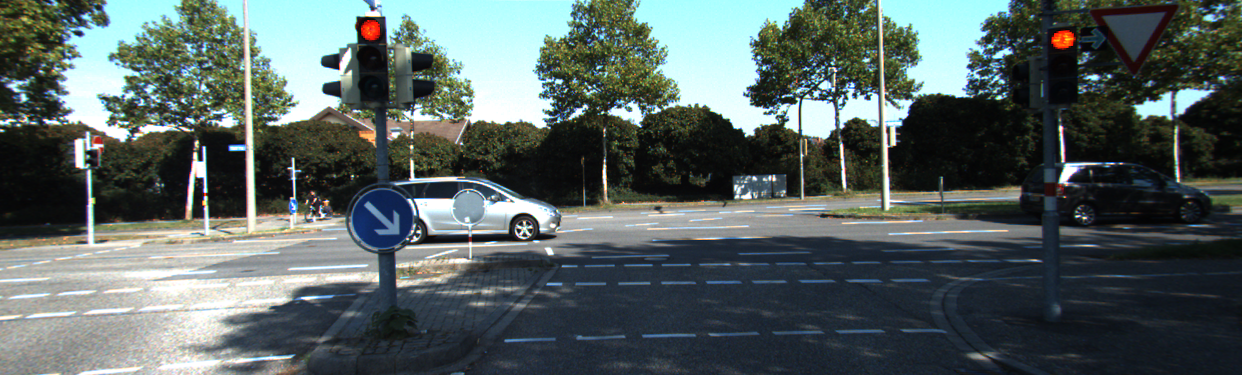}
    &\includegraphics[width=0.325\linewidth]{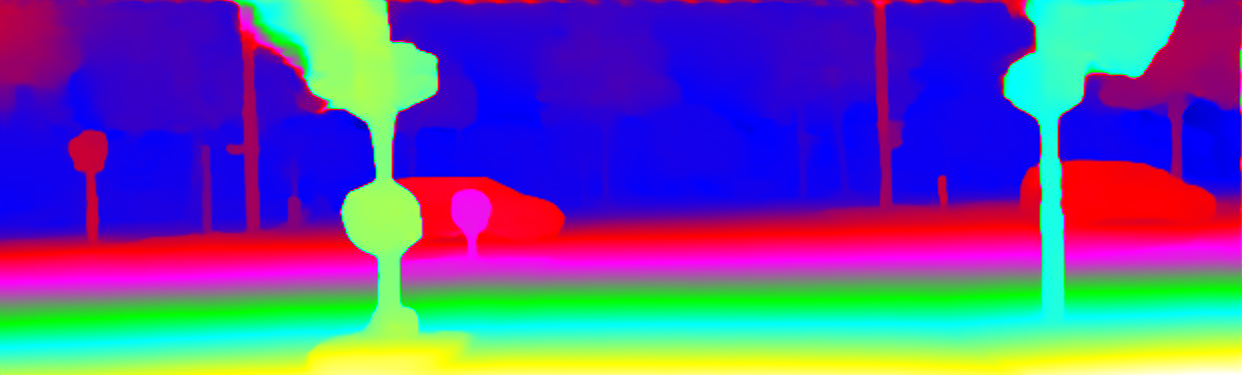}
    &\includegraphics[width=0.325\linewidth]{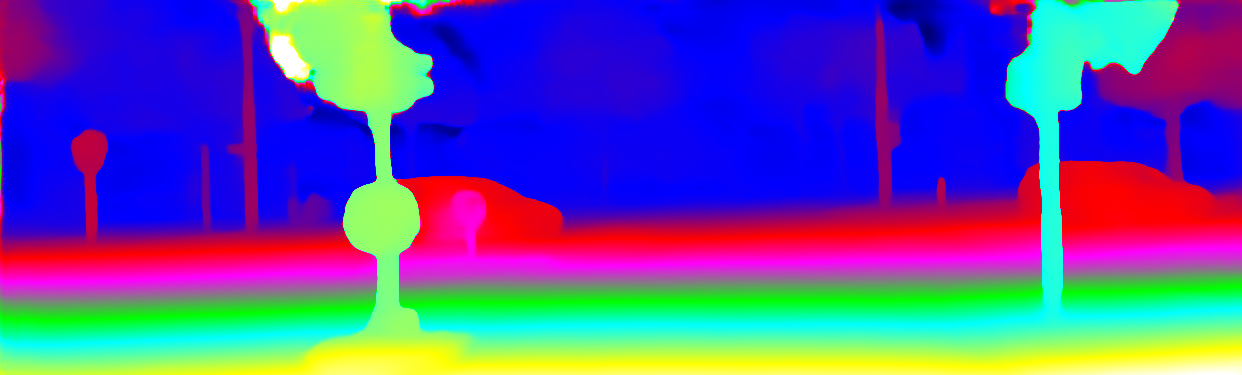}\\
    \end{tabular}
	\caption{\textbf{Qualitative Results on KITTI 2015 \emph{test} dataset.}} 
     \label{fig:kitti}
\end{figure}

\textbf{KITTI 2015.}
Table~\ref{tab:kitti} shows the accuracy and runtime of leading algorithms on the KITTI 2015 benchmark. Our method is the fastest and most accurate among real-time algorithms. Fig.~\ref{fig:kitti} visualizes several results on the test set.

\begin{table}[!t]
\scriptsize
\setlength{\tabcolsep}{1.5mm}
	\centering
	\caption{\textbf{Results on KITTI 2015 \emph{test} set.} Bold indicates the best, while underline indicates the second best.}
	\begin{tabular}{l r c c c c c c c}
\toprule
 & & & \multicolumn{3}{c}{Non-occ (\%)} & \multicolumn{3}{c}{All (\%)} \\
\cmidrule(lr){4-6}
\cmidrule(lr){7-9}
&Method &Runtime(s) &bg &fg &all &bg &fg &all\\
\midrule
\parbox[t]{0.mm}{\multirow{8}{*}{\rotatebox[origin=c]{90}{\scriptsize{Non-real time}}}}

&MC-CNN~\cite{Zbontar2016} &16.00  &2.48 &7.64 &3.33 &2.89 &8.88 &3.89\\
&CRL~\cite{pang2017cascade} &0.26 &2.32 &3.68 &2.36 &2.48 &3.59 &2.67\\
&PSMnet~\cite{chang2018pyramid} &0.31 &1.71 &4.31 &2.14 &1.86 &4.62 &2.32\\
&GC-Net~\cite{KendallMDHKBB17} &0.65 &2.02 &3.12 &2.45 &2.21 &6.16 &2.87\\
&iResNet~\cite{Liang2018Learning} &0.08 &2.15 &2.55 &2.22 &2.35 &3.23 &2.50 \\
&HSM~\cite{Yang_2019_CVPR} &\textbf{0.12} &1.63 &3.40 &1.92 &1.80 &3.85 &2.14\\
&GA-Net-15~\cite{Zhang2019GANet} &0.36 &\textbf{1.40} &3.37 &1.73 &1.55 &3.82 &1.93  \\
&DPruner\_Best~\cite{Duggaliccv2019} &0.13 &1.71 &3.18 &1.95 &1.87 &3.56 &2.15\\
\midrule
\parbox[t]{0.mm}{\multirow{6}{*}{\rotatebox[origin=c]{90}{\scriptsize{Real-time}}}}
&StereoNet~\cite{khamis2018stereonet} &0.02 &- &- &- &4.30 &7.45 &4.83 \\
&MAD-Net~\cite{Tonioni_2019_CVPR} &0.02 &3.45 &8.41 &4.27 &3.75 &9.2 &4.66\\
&DispNetC~\cite{Mayer2016CVPR} &0.04 &4.11 &\textbf{3.72} &4.05 &4.32 &\textbf{4.41} &4.34\\
&DeepCostAggr \cite{Kuzmin2017mlsp} &0.03 &4.82 &10.11 &5.69 &5.34 &11.35 &6.34\\
&RTSNet \cite{Lee2019icip} &0.02 &\underline{2.67} &5.83 &\underline{3.19} &\underline{2.86} &6.19 &\underline{3.41}\\
&Ours &\textbf{0.01} &\textbf{2.12} &\underline{3.88} &\textbf{2.42} &\underline{2.51} &\underline{4.62} &\textbf{2.86}\\
\bottomrule
	\end{tabular}
\label{tab:kitti}
\end{table}

\begin{figure}[t]
    \centering
    \tabcolsep=0.01cm
    \begin{tabular}{c c c c}
    \scriptsize RGB & \scriptsize DeepPruner & \scriptsize HSM & \scriptsize Ours\\
    \includegraphics[width=0.245\linewidth]{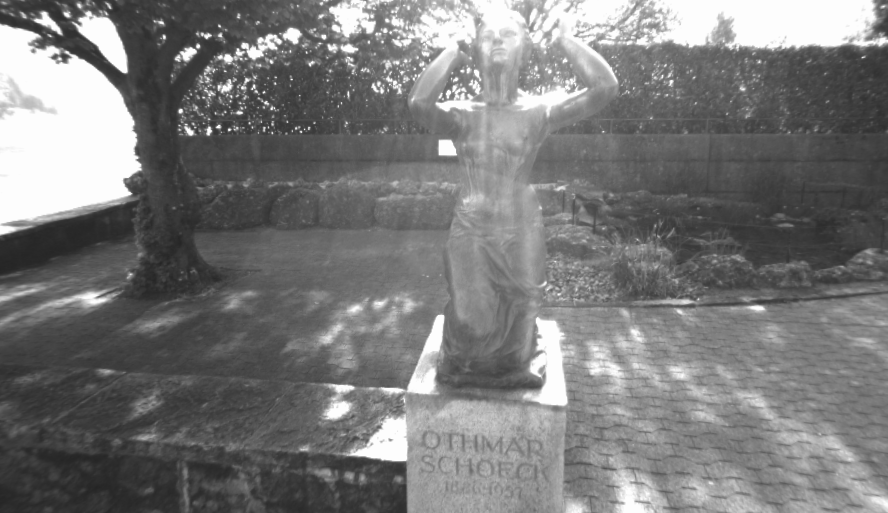}
    &\includegraphics[width=0.245\linewidth]{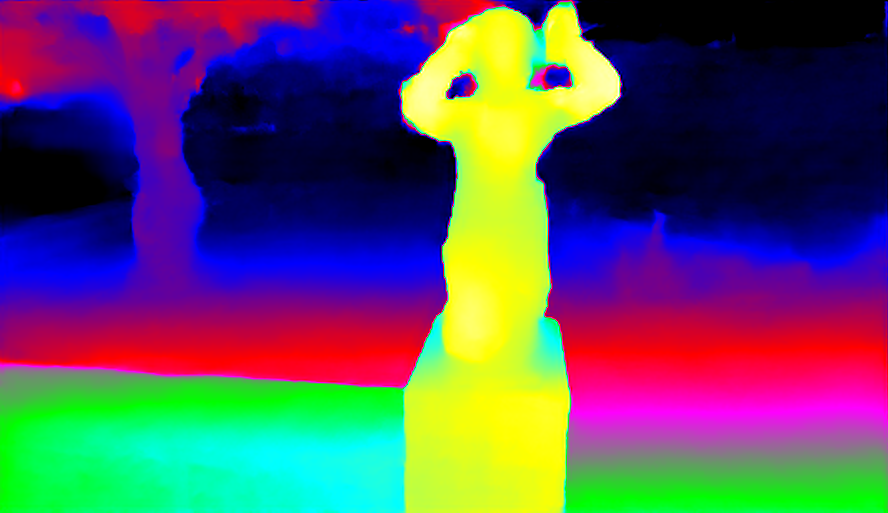} 
    &\includegraphics[width=0.245\linewidth]{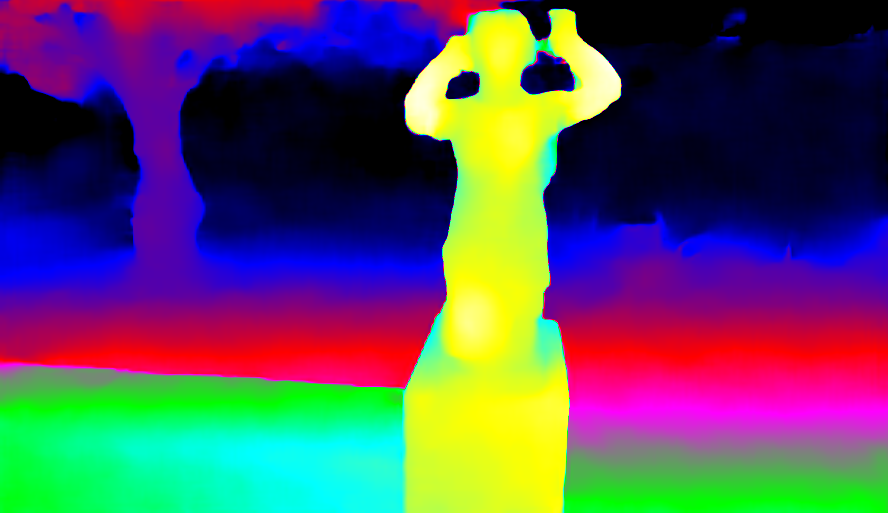}
    &\includegraphics[width=0.245\linewidth]{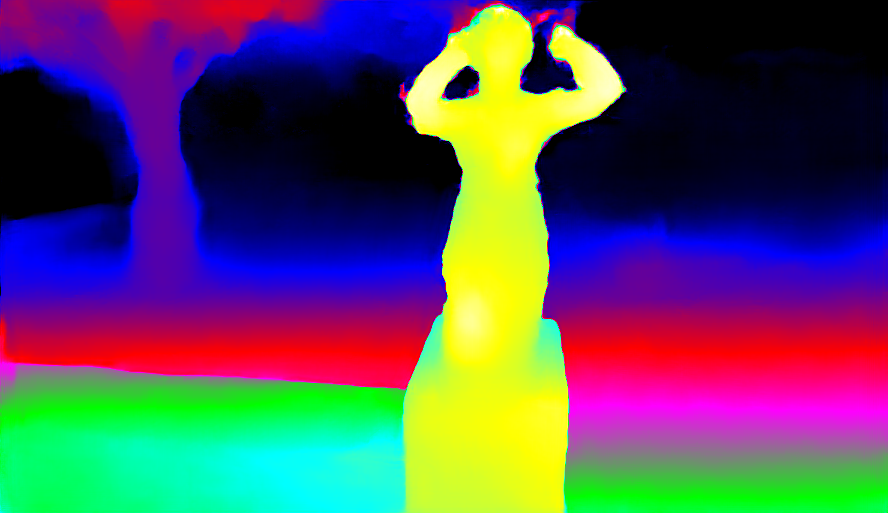}\\
    \includegraphics[width=0.245\linewidth]{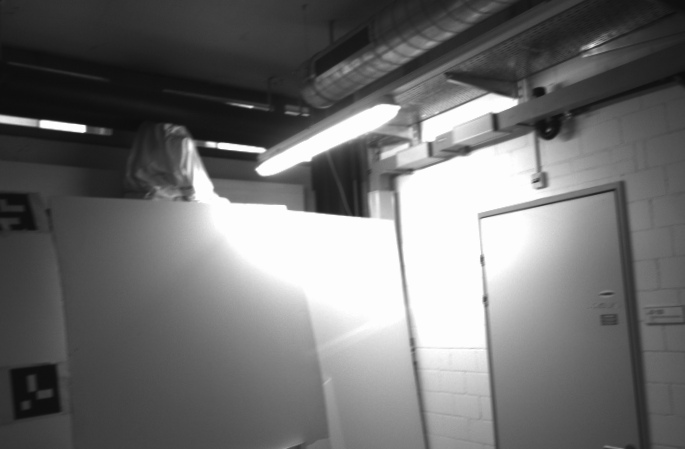}
    &\includegraphics[width=0.245\linewidth]{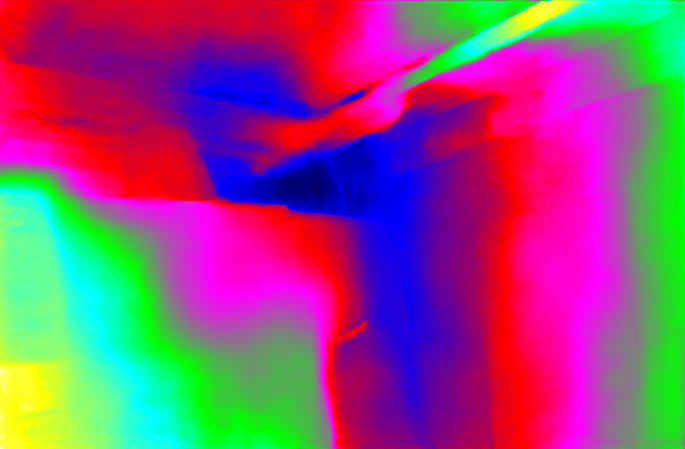} 
    &\includegraphics[width=0.245\linewidth]{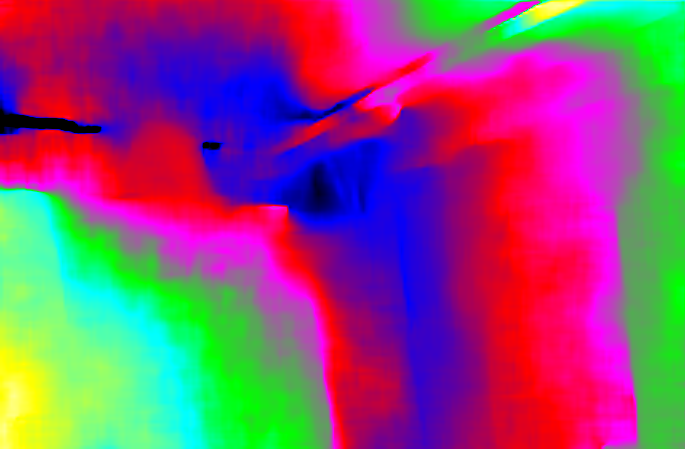}
    &\includegraphics[width=0.245\linewidth]{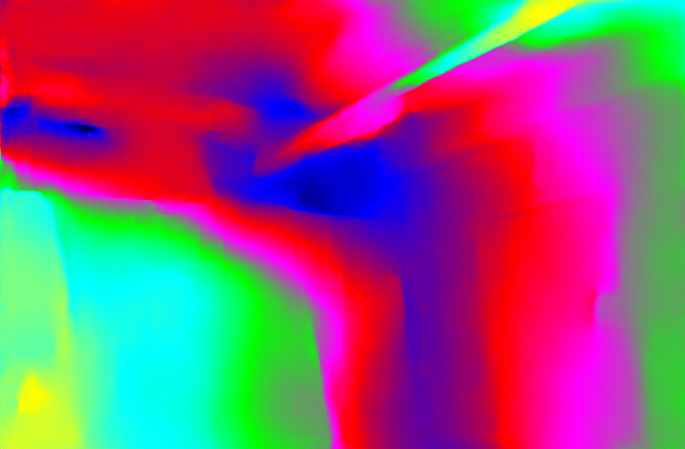}\\
    \end{tabular}
	\caption{\textbf{Qualitative results on ETH3D \emph{test} dataset.}}
    \label{fig:eth3d}
\end{figure}

\begin{table}[!t]
\scriptsize
\centering
\caption{\textbf{Results on ETH3D \emph{test} dataset}. Bold indicates the best, while underline indicates the second best.}
\tabcolsep=0.15cm
\begin{tabular}{c c c c c c c c}
\toprule
Methods & time(s) & EPE  & rmse & bad-4.0 & bad-2.0 & bad-1.0  & A99\\ 
\hline
HSM \cite{Yang_2019_CVPR}  &\textbf{0.12} &\underline{0.29} &0.67 &0.68 &1.48  &4.25  &3.25                          \\ 
SDRNet \cite{Chabra_2019_CVPR} &0.13 &0.34 &0.71 &0.50 &1.66   &6.02  &3.07                     \\ 
iResNet \cite{Liang2018Learning} &0.15 &0.25 &\underline{0.59} &\textbf{0.34}  &\underline{1.20} &\underline{4.04}  &\underline{2.70}                          \\ 
DPruner \cite{Duggaliccv2019} &0.12 &\textbf{0.28} &\textbf{0.58}  &\textbf{0.34}   &\textbf{1.04} &\textbf{3.82}  &\textbf{2.61}                     \\
PSMnet \cite{chang2018pyramid} &0.31 &0.36 &0.75  &0.54 &1.31  &5.41 &3.38                         \\
\hline
DN-CSS \cite{DNCSS2018} &0.05  &\textbf{0.24} &\textbf{0.56}   &\textbf{0.38}  &\textbf{0.96}  &\textbf{3.00}  &\underline{2.89}                       \\ 
Ours & \textbf{0.01} &\underline{0.32} &\underline{0.63} &\underline{0.53} &\underline{1.25} &\underline{4.82} &\textbf{2.79}                                            \\
\bottomrule
\end{tabular}
\label{tab:eth3d}
\end{table}

\begin{figure}[t]
    \centering
    \tabcolsep=0.01cm
    \begin{tabular}{c c c c}
    \scriptsize RGB & \scriptsize DeepPruner & \scriptsize HSM & \scriptsize Ours\\
    \includegraphics[width=0.245\linewidth]{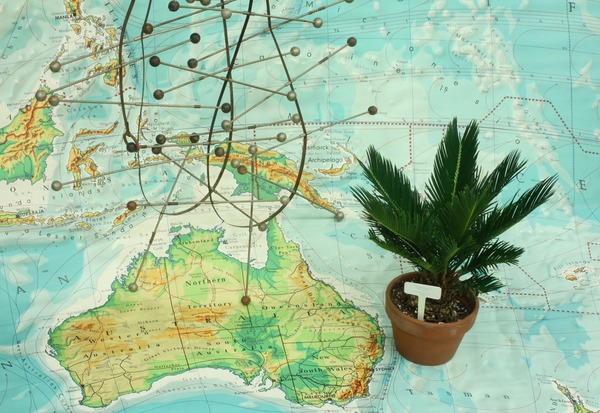}
    &\includegraphics[width=0.245\linewidth]{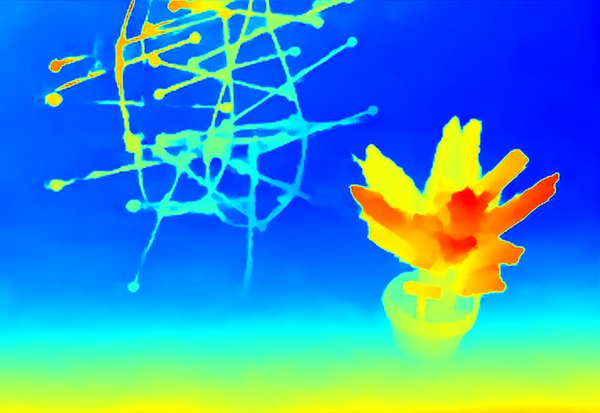} 
    &\includegraphics[width=0.245\linewidth]{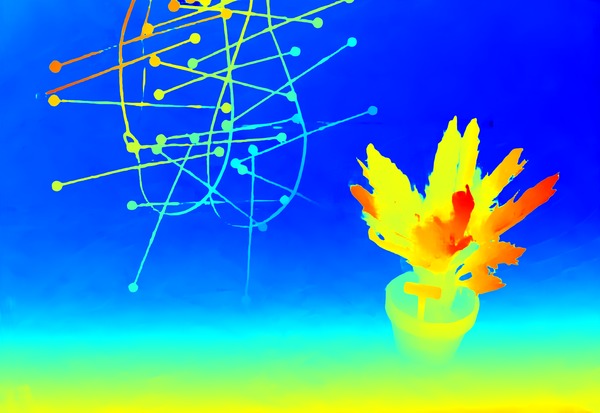}
    &\includegraphics[width=0.245\linewidth]{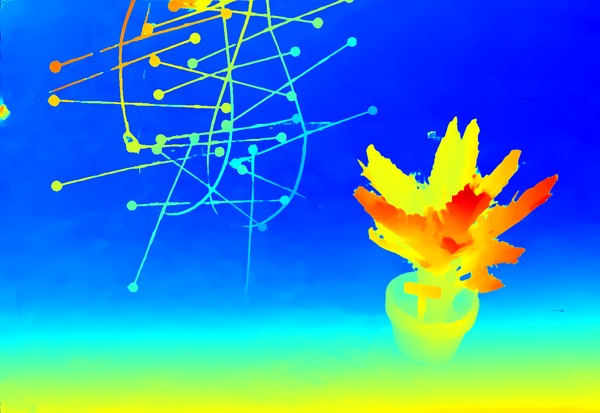} \\
    \includegraphics[width=0.245\linewidth]{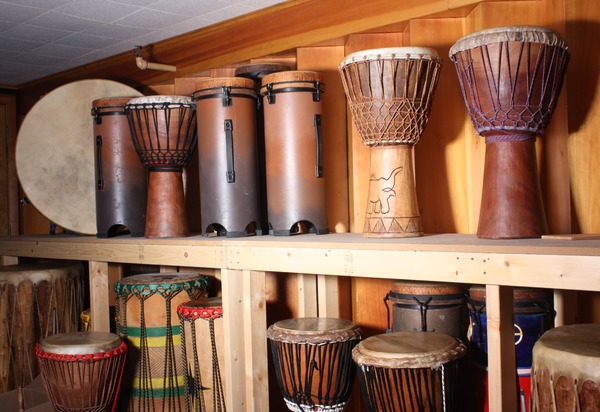}
    &\includegraphics[width=0.245\linewidth]{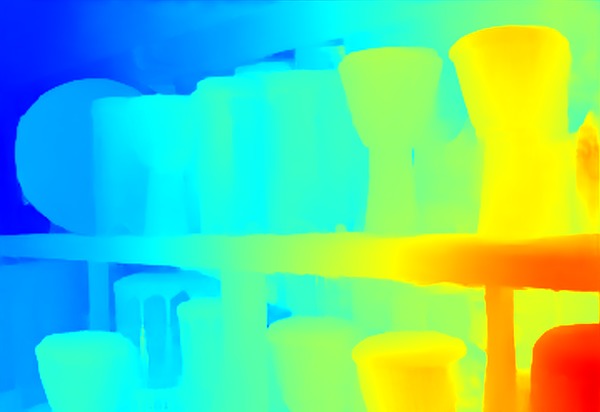} 
    &\includegraphics[width=0.245\linewidth]{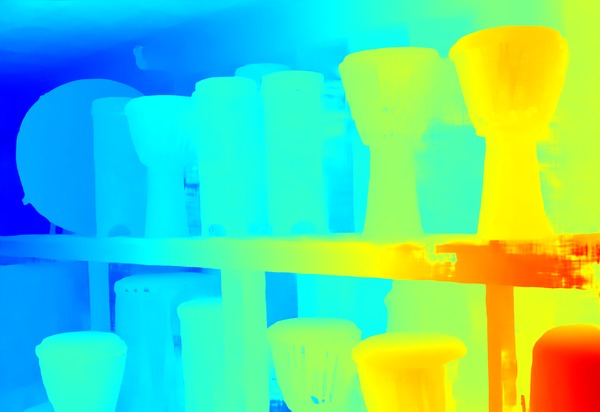}
    &\includegraphics[width=0.245\linewidth]{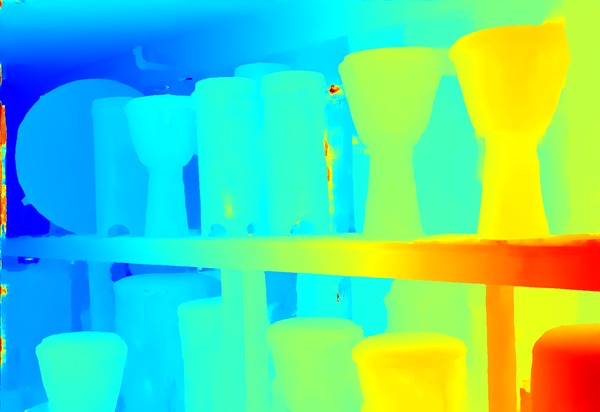} \\
    \end{tabular}
	\caption{\textbf{Qualitative results on Middlebury 2014  \emph{test} dataset.}}
    \label{fig:mb}
\end{figure}
\begin{table}[!t]
\scriptsize
\centering
\caption{\textbf{Results on Middlebury 2014  \emph{test} dataset}. Bold indicates the best, while underline indicates the second best.}
\tabcolsep=0.15cm
\begin{tabular}{c c c c c c c c}
\toprule
Methods & time(s) & EPE  & rmse & bad-4.0 & bad-2.0 & bad-1.0  & A99\\ 
\hline
SGM \cite{Hirschmuller2008}    &0.32 &5.32   &20.0  &12.2   &18.4    &\underline{31.1}	   &109	              \\ 
HSM \cite{Yang_2019_CVPR}      &0.47 &\textbf{2.07}   &\textbf{10.3}  &\textbf{4.83}   &\textbf{10.2}	 &\textbf{24.6}     &\textbf{39.2}         \\ 
iResNet \cite{Liang2018Learning}&0.28 &3.31	 &\underline{11.3}	&12.6   &22.9    &38.8     &\underline{48.6}          \\ 
DPruner \cite{Duggaliccv2019}  &0.11 &4.80   &14.7	&15.9   &30.1    &52.3     &67.7    \\
PSMNet \cite{chang2018pyramid} &0.45 &6.68	 &19.4  &23.5   &42.1	 &63.9     &84.5    \\
DN-CSS \cite{DNCSS2018}        &0.58 &4.04   &13.9  &14.7	&22.8    &36.0     &58.8     \\ 
\hline
Ours                          &\textbf{0.04} &\underline{3.12}   &13.8  &\underline{7.22}   &\underline{15.4}	 &35.1	   &55.6       \\
\bottomrule
\end{tabular}
\label{tab:mb}
\end{table}

\textbf{ETH3D.}
Table~\ref{tab:eth3d} shows quantitative results of our method on the ETH3D benchmark. Our method achieves competitive performance while being significantly faster than all the other methods. A qualitative comparison of our method with other SOTA algorithms is shown in Fig.~\ref{fig:eth3d}.

\textbf{Middlebury 2014.}
Our method is the only deep learning-based method that can achieve real-time performance on the Middlebury dataset. By accuracy, it is the second best among all competitors, as shown in Table~\ref{tab:mb}. Compared to the most accurate approach (HSM), our method is 12$\times$ faster while has comparable accuracy. Qualitative results are shown in Fig.~\ref{fig:mb}.

\subsection{Model Design Analysis}
\label{sec:ab}
In this section, we analyze the components of our network structure and justify the design choices, including the performance differences between our proposed method, hand-crafted cost computation and  Baseline3D, the effectiveness of entropy map and the robustness and generalizability of our network. All experiments are conducted on the SceneFlow dataset except the robustness experiment.
\begin{table}[!t]
 \centering
\scriptsize
\setlength{\tabcolsep}{1.5mm}
\centering
\caption{\textbf{Contributions of each component.} The first 3 rows compare the performance (both speed and accuracy) between hand-crafted cost, Baseline3D and our method. The last 4 rows show the contribution of each components of our method.}
\begin{tabular}{c c c c c c c c c}
\toprule
\multicolumn{6}{c}{Architecture Variant} & Inference &\multicolumn{2}{c}{SceneFlow} \\
Dot & 3D & 2D & RefineNet &Entropy &Warp &time(s) &EPE & bad-1.0\\
\hline
\checkmark & & & & & &0.001 &6.05 &59.13\\
\hline
& \checkmark & & & &&0.15  &0.78 &8.11\\
\hline
& &\checkmark & & &&0.007 &1.20 &10.31\\
& &\checkmark &\checkmark & &&0.01  &1.14 &10.16\\
& &\checkmark &\checkmark & &\checkmark &0.02  &1.08 &9.84\\
& &\checkmark &\checkmark &\checkmark & &0.01  &1.09 &9.67\\
\bottomrule
\end{tabular}
\label{tab:arch}
\end{table}

To fairly analyze the benefits and disadvantages of our method and volumetric method (Baseline3D), we compare them using exactly the same training strategy, and settings. 
For the sake of completeness, we also compare the performance with hand-crafted cost using $c_{SSD}(\mathbf{p},d) =  \sum_{\mathbf{q}\in \mathcal{N}_\mathbf{p}}\left\|f(I^L(\mathbf{p}))- f(I^R(\mathbf{p}-d)) \right \|_2^2,$ where $\mathcal{N}_\mathbf{p}$ is a local patch around $\mathbf{p}$.

A detailed comparison is shown in Table.~\ref{tab:arch}. As expected, using hand-crafted costs fails to generate meaningful results. Baseline3D achieves the best performance in accuracy but is 15$\times$ slower than Ours. Compared with Baseline3D, the performance of proposed method drops from 0.78 pixels EPE to 1.09 pixels, which is acceptable in many real-world scenarios. In fact, our method achieves the top accuracy among all real-time algorithms as shown in Table.~\ref{tab:2dvs3dsceneflow}. 

\begin{figure}[t]
    \centering
    \tabcolsep=0.03cm
    \begin{tabular}{c c c c c}
   \scriptsize RGB & \scriptsize Baseline3D & \scriptsize Ours &  \scriptsize Baseline3D err &\scriptsize Ours err\\
    \includegraphics[width=0.195\linewidth]{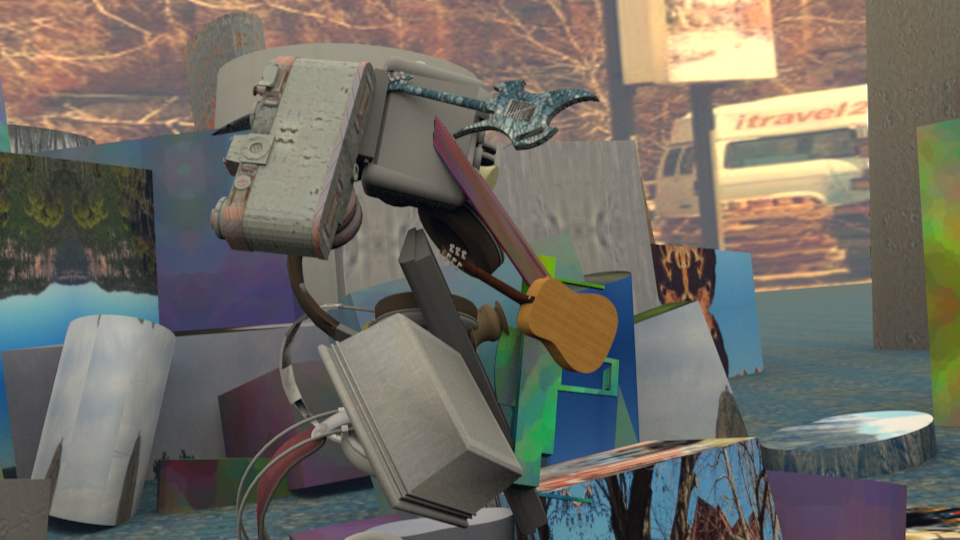}
    &\includegraphics[width=0.195\linewidth]{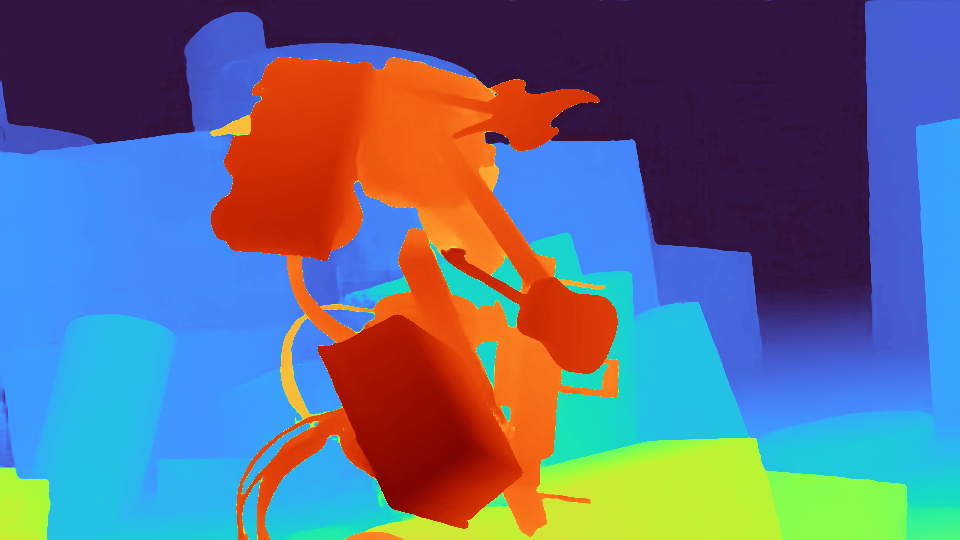}
    &\includegraphics[width=0.195\linewidth]{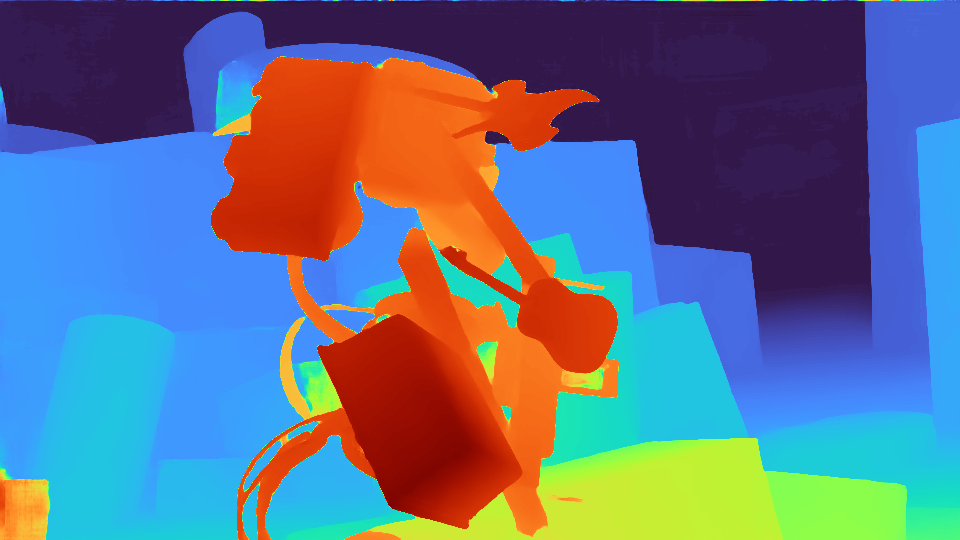}
    &\includegraphics[width=0.195\linewidth]{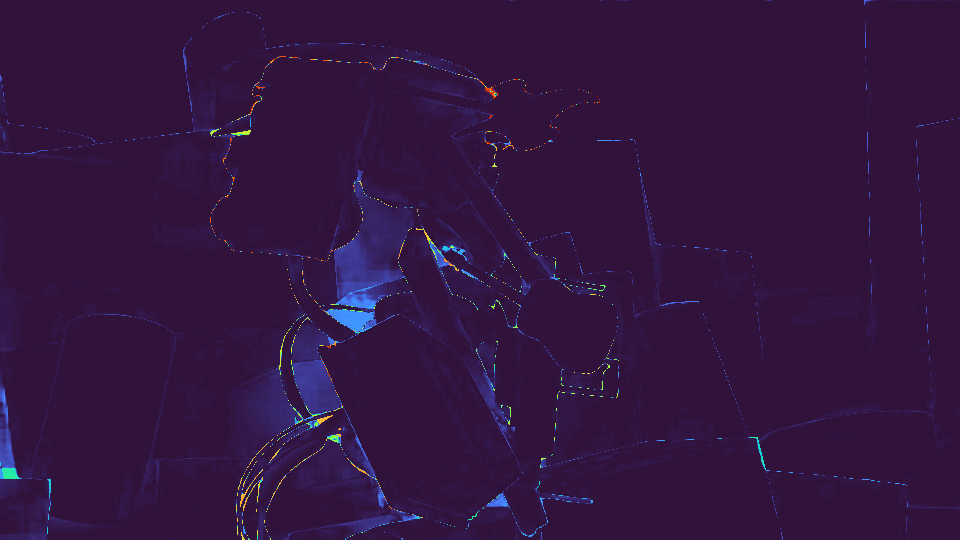}
    &\includegraphics[width=0.195\linewidth]{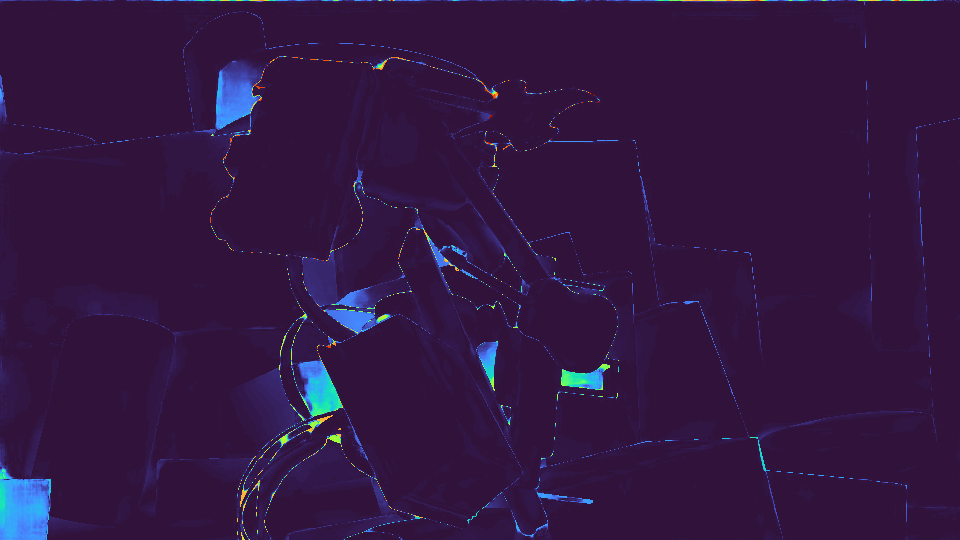} \\
    \includegraphics[width=0.195\linewidth]{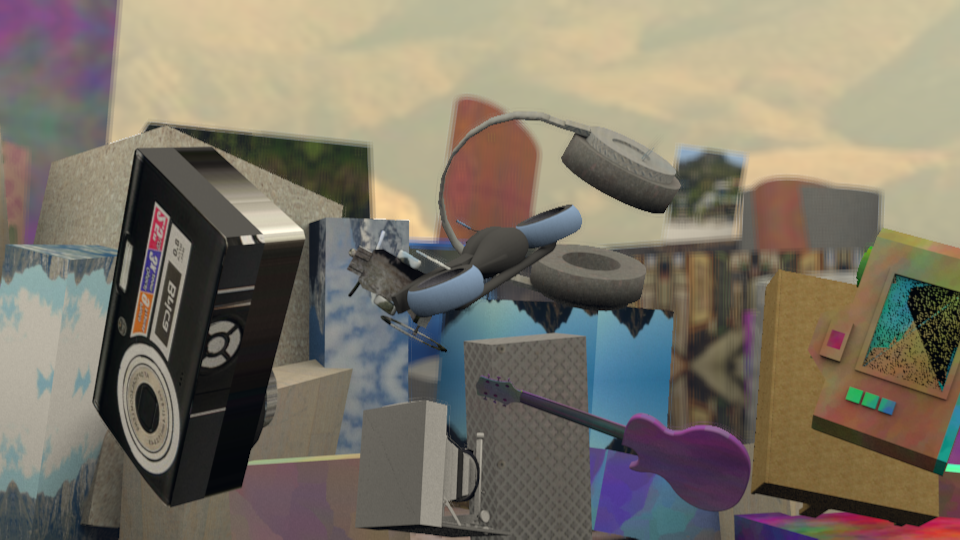}
    &\includegraphics[width=0.195\linewidth]{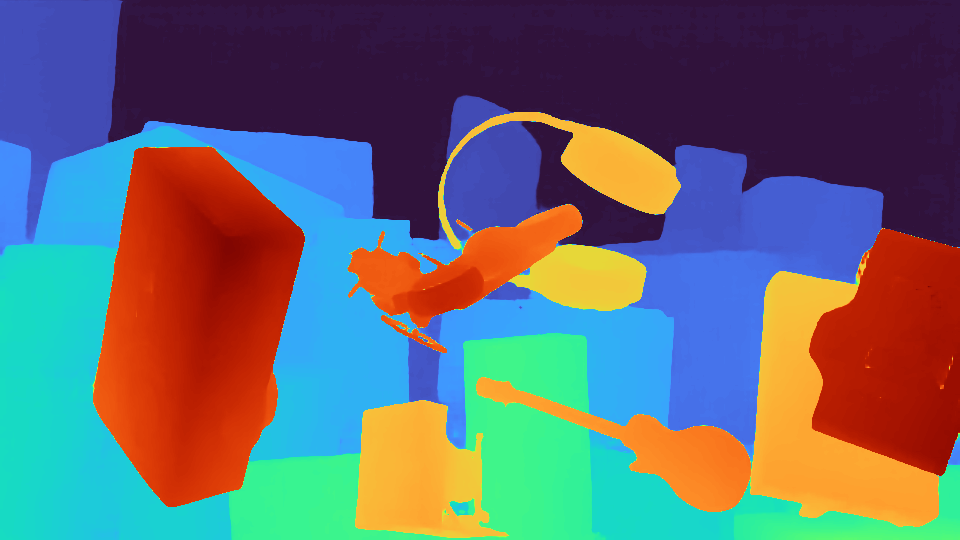}
    &\includegraphics[width=0.195\linewidth]{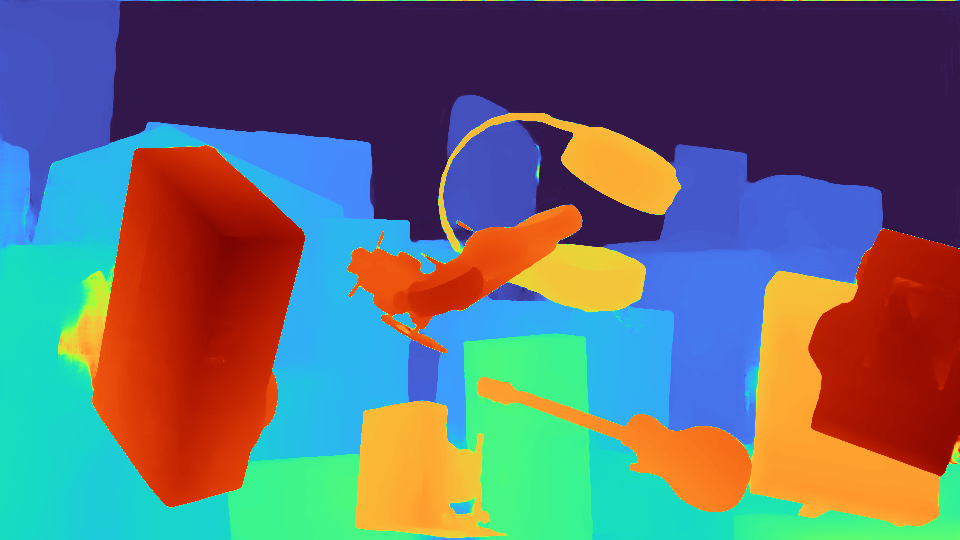}
    &\includegraphics[width=0.195\linewidth]{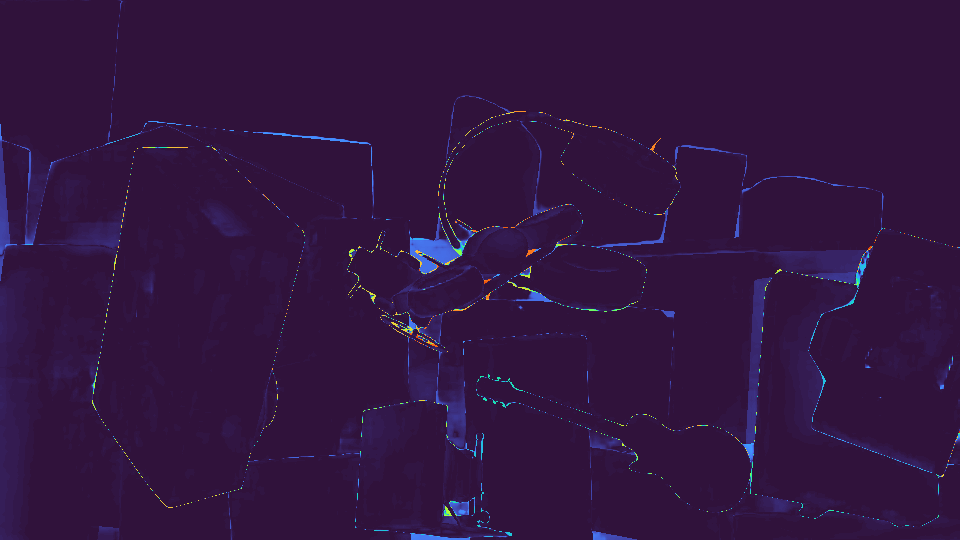}
    &\includegraphics[width=0.195\linewidth]{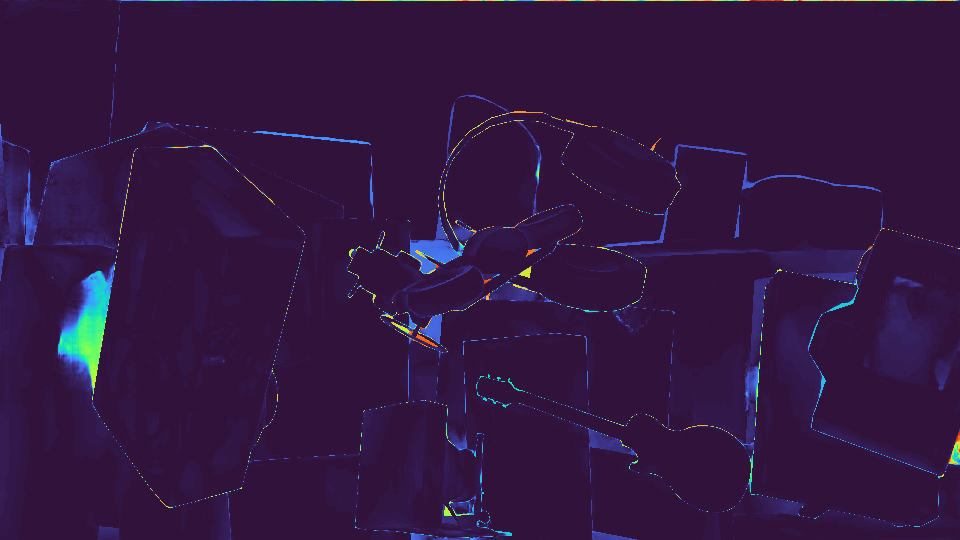}\\
    \end{tabular}
	\caption{\textbf{Qualitative comparison on the SceneFlow dataset.}}
    \label{fig:2dvs3d}
\end{figure}

A qualitative comparison of Ours and Baseline3D is provided in Fig.~\ref{fig:2dvs3d} and Fig.~\ref{fig:kitti}, where both approaches generate high-quality disparity maps. Baseline3D has better occlusion handling ability: By jointly regularizing spatial and disparity domain, the network can better pre-scale the matching costs to handle the multi-modal situation. However, Baseline3D has a drawback that we will address in the next section. 
We further analyze the contribution of each component of proposed network, including the use of refine net and the entropy of the cost volume as input to the refine net. In Table.~\ref{tab:arch}, we can see that by adding the refine net, the EPE drops from 1.20 to 1.14 and the bad-1.0 drops from 10.31 to 10.16. Using a warped right image can further boosts the accuracy more than 5\% but significantly increases the processing time, whereas adding the entropy map can have similar performance without sacrificing the efficiency. We posit that the entropy map provides evidence for the refine net as to which parts of the disparity map are unreliable and need to be smoothed using priors. 
\begin{figure}[t]
    \centering
    \tabcolsep=0.03cm
    \begin{tabular}{c c c c c}
 \scriptsize RGB & \scriptsize Baseline3D & \scriptsize Ours & \scriptsize Baseline3D err &\scriptsize Ours err\\
    \includegraphics[width=0.195\linewidth]{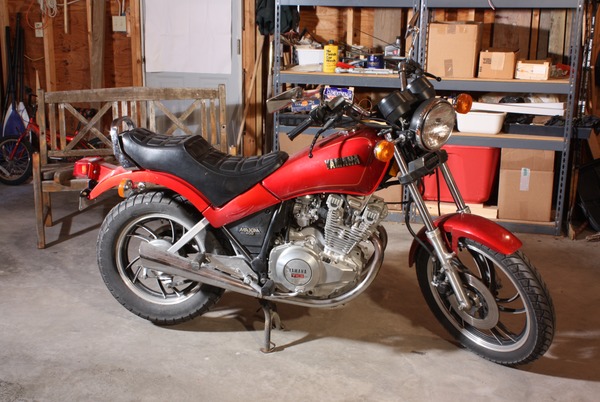}
    &\includegraphics[width=0.195\linewidth]{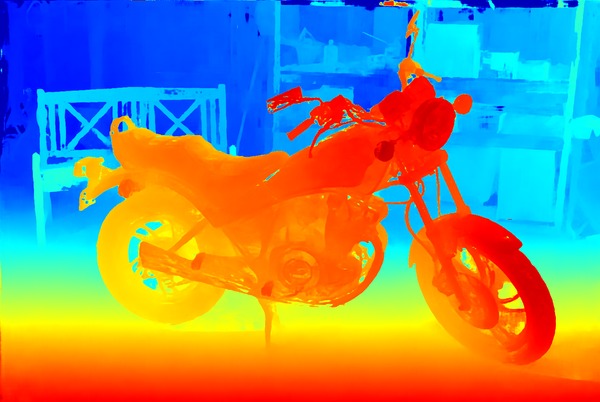}
    &\includegraphics[width=0.195\linewidth]{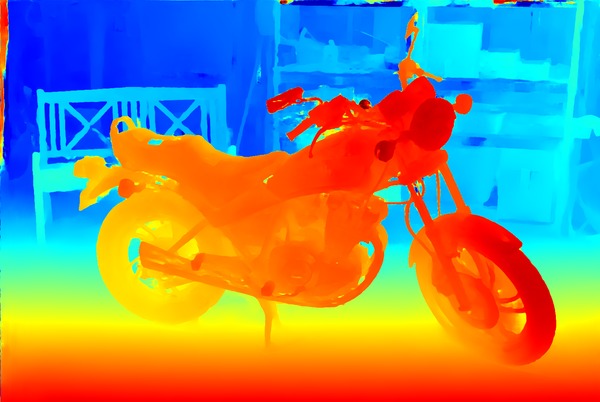}
    &\includegraphics[width=0.195\linewidth]{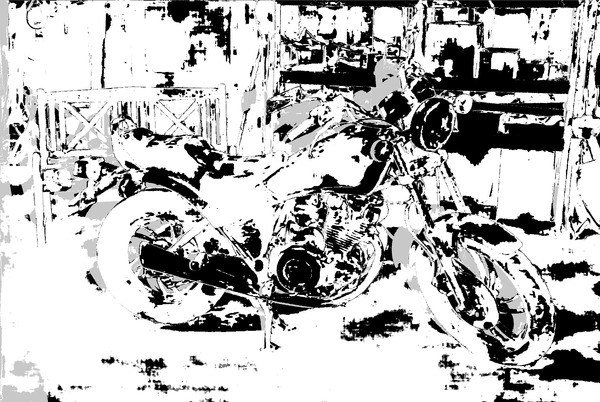}
    &\includegraphics[width=0.195\linewidth]{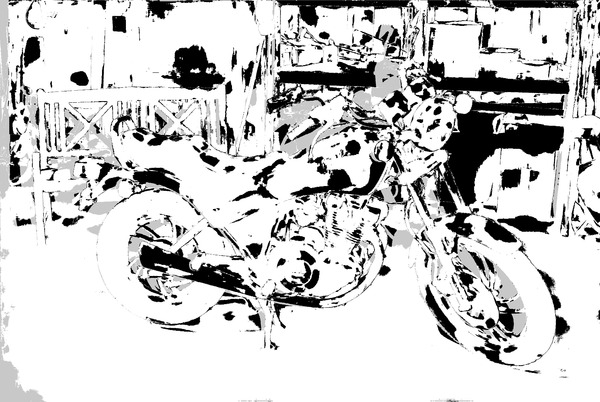}\\
    \includegraphics[width=0.195\linewidth]{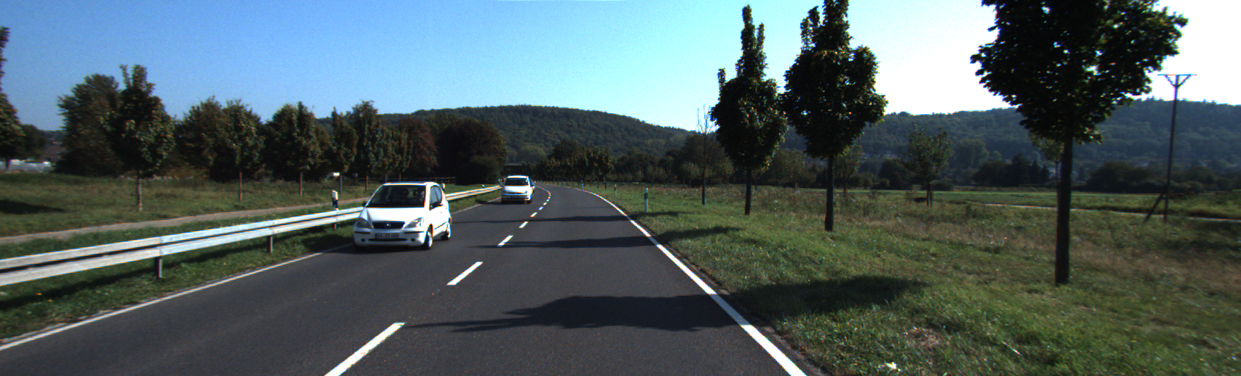}
    &\includegraphics[width=0.195\linewidth]{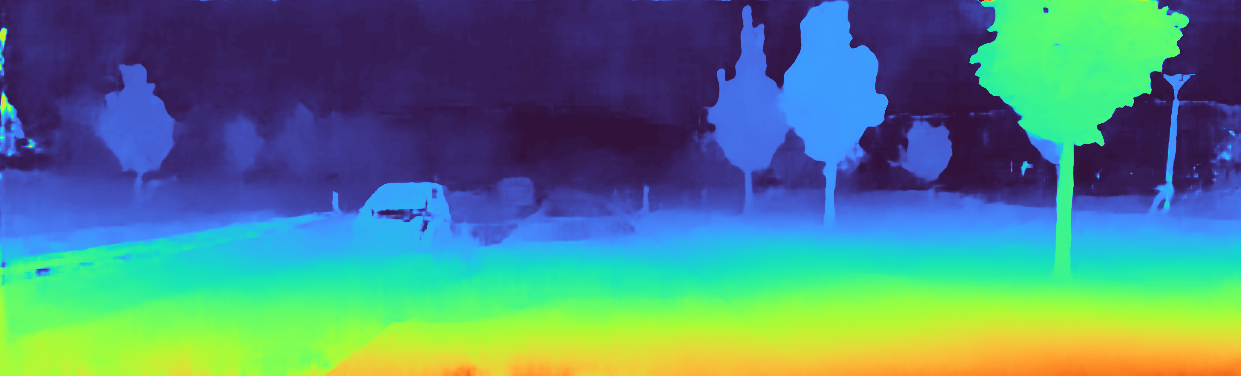}
    &\includegraphics[width=0.195\linewidth]{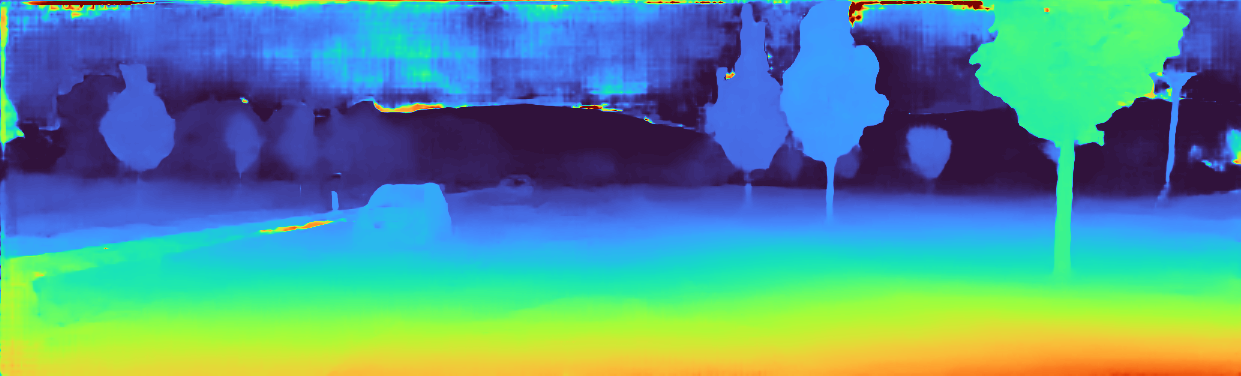}
    &\includegraphics[width=0.195\linewidth]{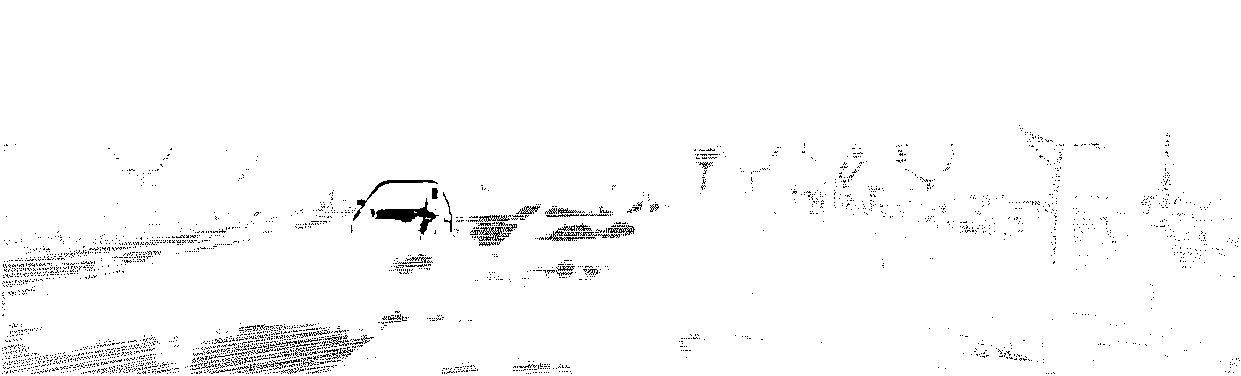}
    &\includegraphics[width=0.195\linewidth]{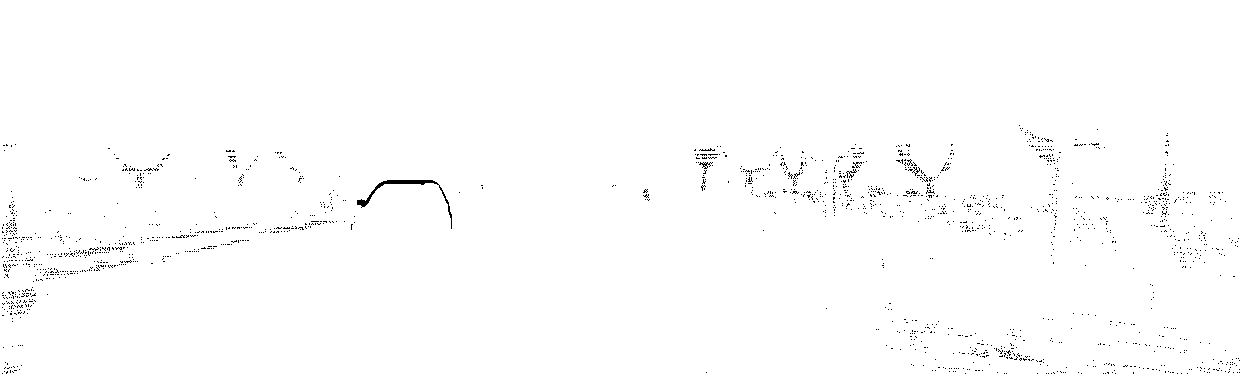}\\
    \end{tabular}
	\caption{\textbf{Qualitative comparison on cross-dataset study.} The SceneFlow pre-trained models were tested on Middlebury 2014 and KITTI 2015 training sets \emph{without any finetuning.}  Note that Ours generalizes better across datasets than Baseline3D.}
    \label{fig:crossdatasets}
\end{figure}

\begin{table}[!t]
\scriptsize
\centering
\caption{\textbf{Quantitative results on SceneFlow dataset}. Our model is the fastest and the most accurate among real-time algorithms.}
\tabcolsep=0.2cm
\begin{tabular}{l r c c c c}
\toprule
&Methods & EPE (pix) & bad-1.0 & Params & Runtime(s) \\ 
\midrule
\parbox[t]{2mm}{\multirow{6}{*}{\rotatebox[origin=c]{90}{\scriptsize{Non-real time}}}}
&GC-Net\cite{KendallMDHKBB17} & 1.84 & 15.6 & 3.5M & 0.65 \\
&CRL\cite{pang2017cascade} & 1.32 & - & 78.77M & 0.26 \\
&PSMNet\cite{chang2018pyramid} & 1.09 & 12.1 & 5.22M & 0.31 \\
&GA-Net\cite{Zhang2019GANet} &{\bf0.78} & 8.7 &6.58M &1.8 \\
&DPruner\cite{Duggaliccv2019} &0.86 & - & - & 0.13 \\
&Baseline3D      & {\bf0.78}    & {\bf8.11}  &4.26M  & {\bf0.15} \\ 
\midrule
\parbox[t]{2mm}{\multirow{2}{*}{\rotatebox[origin=c]{90}{\tiny{Realtime}}}}
&DispNetC\cite{Mayer2016CVPR} &1.68 & - & - & 0.04 \\
&StereoNet\cite{khamis2018stereonet}& 1.10 & - &- & 0.02\\
&Ours  & {\bf 1.09}   & {\bf 9.67}    &1.70M & {\bf 0.01}  \\
\bottomrule
\end{tabular}
\label{tab:2dvs3dsceneflow}
\end{table}

In Table~\ref{tab:2dvs3dsceneflow}, we compare our methods with SOTA deep stereo methods on the SceneFlow dataset. Baseline3D achieves top performance (in terms of both accuracy and speed) among non-real time methods, while our method with refinement achieves the top performance among real-time methods. It is worth noting that previous 2D convolution based methods \cite{pang2017cascade,Liang2018Learning} need a large number of parameters to learn the context mapping between inputs and disparity maps. By using proposed cost computation, our network achieves better accuracy, is significantly faster, and requires a fraction of the parameters as previous 2D convolution based methods.

\subsection{Robustness and Generalizability}
In the previous section, we showed that Baseline3D yields higher accuracy than Ours, while being significantly slower. Aside from speed and memory consumption, we consider other drawbacks in using 3D convolutions.
Since spatial information should be independent of disparity, there is a risk that using 3D convolutions may cause the network to learn relationships that are present only in one specific dataset, which may actually hinder its generalization performance on other unseen datasets.

To test this hypothesis, we compared the generalizability of Ours and Baseline3D methods using the SceneFlow-pretrained models, by testing them on the Middlebury and KITTI 2015 training sets, without fine-tuning. Table~\ref{tab:crossdatasets} shows the quantitative results. For better comparison, we use SOTA method GA-Net \cite{Zhang2019GANet} as the reference method. Ours consistently achieves better cross-dataset performance on these two datasets, supporting our hypothesis. The results in Fig.~\ref{fig:crossdatasets} show that Baseline3D generates false boundaries on the road plane while Ours does not, because the former has learned spurious associations from the training set. 

\begin{table}[t]
\scriptsize
\setlength{\tabcolsep}{0.5mm}
	\centering
	\caption{\textbf{Results on cross-dataset study.} SceneFlow pre-trained models were tested on Middlebury 2014 and KITTI 2015 training sets \emph{directly.}}
	\begin{tabular}{l c c c c c c c c c}
\toprule
 & \multicolumn{5}{c}{Middlebury} & \multicolumn{4}{c}{KITTI 2015} \\
\cmidrule(lr){2-6}
\cmidrule(lr){7-10}
Methods &bad-1.0 &bad-2.0 &bad-4.0 &avgerr &rms &EPE &bad-1.0 &bad-2.0 &bad-3.0\\
\midrule
GA-Net \cite{Zhang2019GANet} &59.9 &38.1 &19.8 &\textbf{4.94} &\textbf{13.6} &1.70 &42.35\% & 18.29\%  &\textbf{10.77}\%\\
\hline
Baseline3D &53.0 &33.7 &20.1 &13.5 &34.4 &1.92 &46.76\% &21.02\% &12.03\%\\
Ours &\textbf{51.4} &\textbf{33.1} &\textbf{19.6} &6.67 &19.2 &\textbf{1.63} &\textbf{37.55}\% &\textbf{17.54}\% &10.88\%\\
\bottomrule
	\end{tabular}
\label{tab:crossdatasets}
\end{table}

\subsubsection{Random Dot Stereo}
Random Dot Stereograms (RDSs) \cite{Bela1971} were introduced many decades ago to evaluate depth perception in the human visual system. 
RDSs were key to establishing the fact that the human visual system is capable of estimating depth simply by matching patterns in the left and right images, without utilizing any monocular cues (lighting, color, texture, shape, familiar objects, and so forth).
Similarly, we argue that artificial stereo algorithms should be able to process random dot stereograms, and that this ability provides key evidence to understand the actual behavior of the networks, \ie context mapping or matching. 

In random dot stereo pairs, there is no semantic context. Therefore, methods that fail this test, but that otherwise do well on benchmarks, are mostly likely relying on monocular, semantic cues for inference.
Such methods (including DispNet, and those that build upon DispNet) will struggle to process images whose distribution varies significantly from the training distribution, because they do not actually match pixels between images. 

\begin{figure}[!t]
    \centering
    \tabcolsep=0.03cm
    \begin{tabular}{c c c c}
  \small Left Image & \small Ground Truth & \small Ours & \small MADNet \cite{Tonioni_2019_CVPR} \\
    \includegraphics[width=0.245\linewidth]{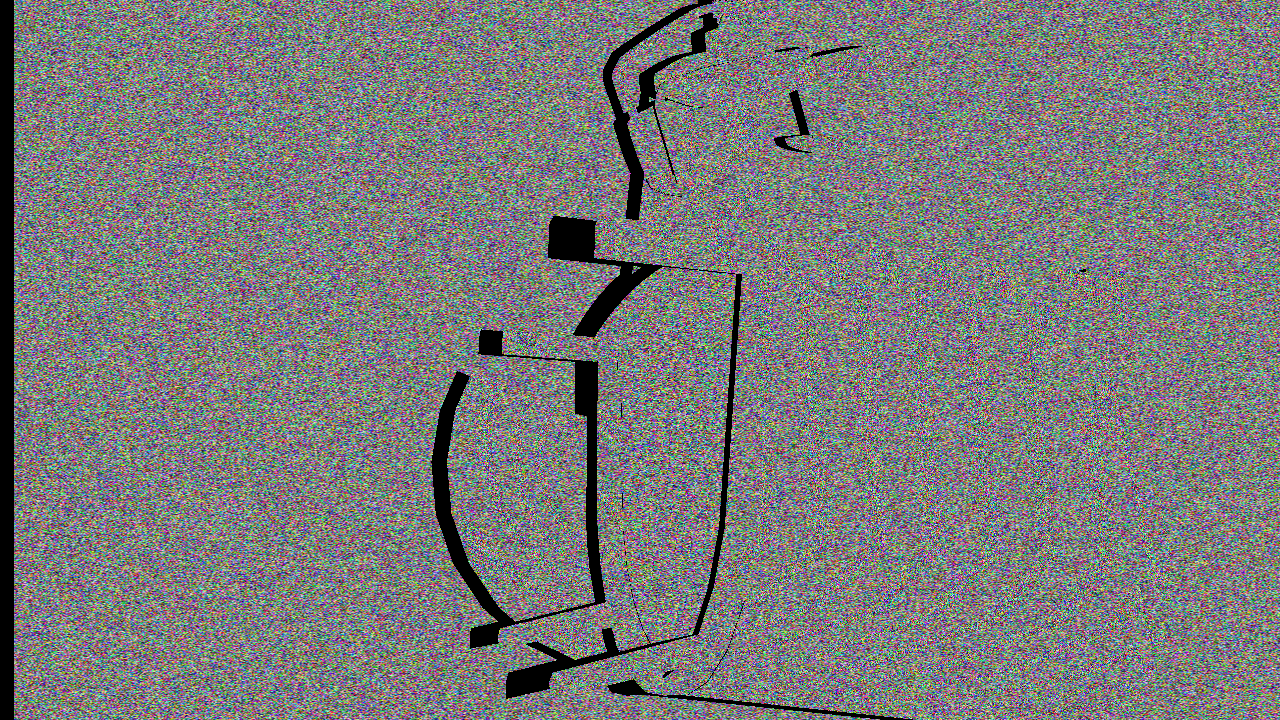}
    & \includegraphics[width=0.245\linewidth]{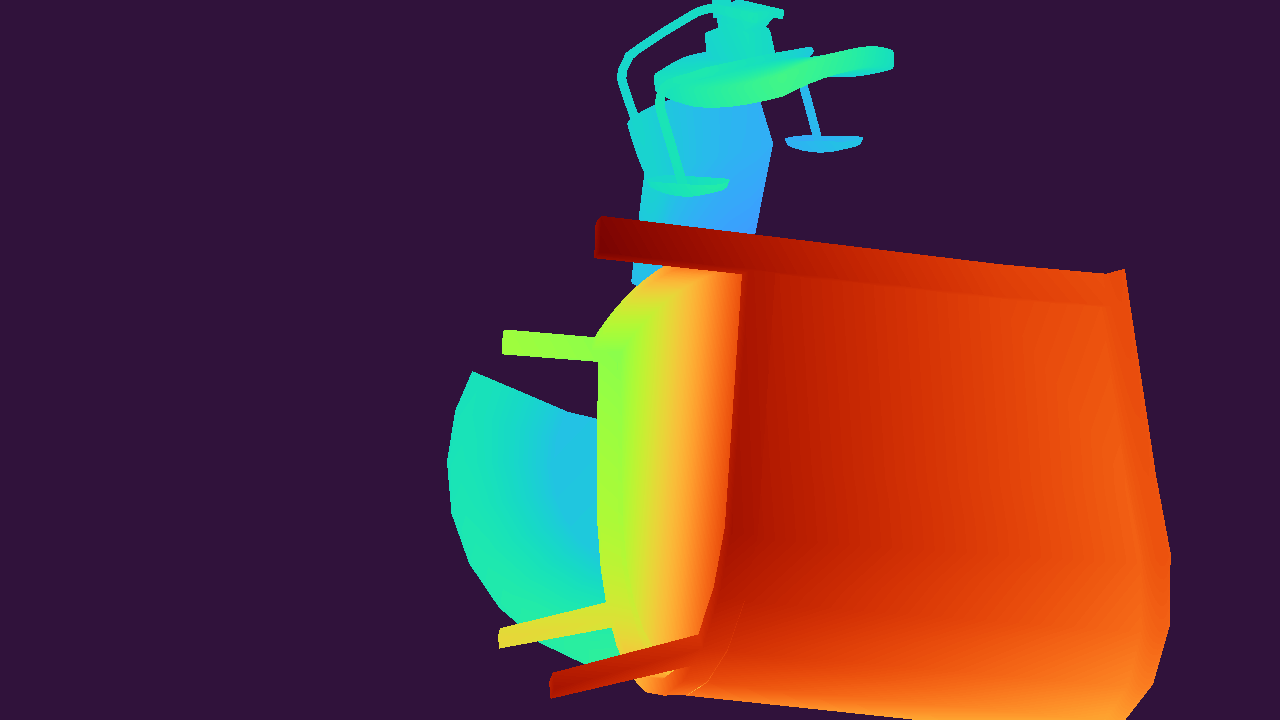}
    & \includegraphics[width=0.245\linewidth]{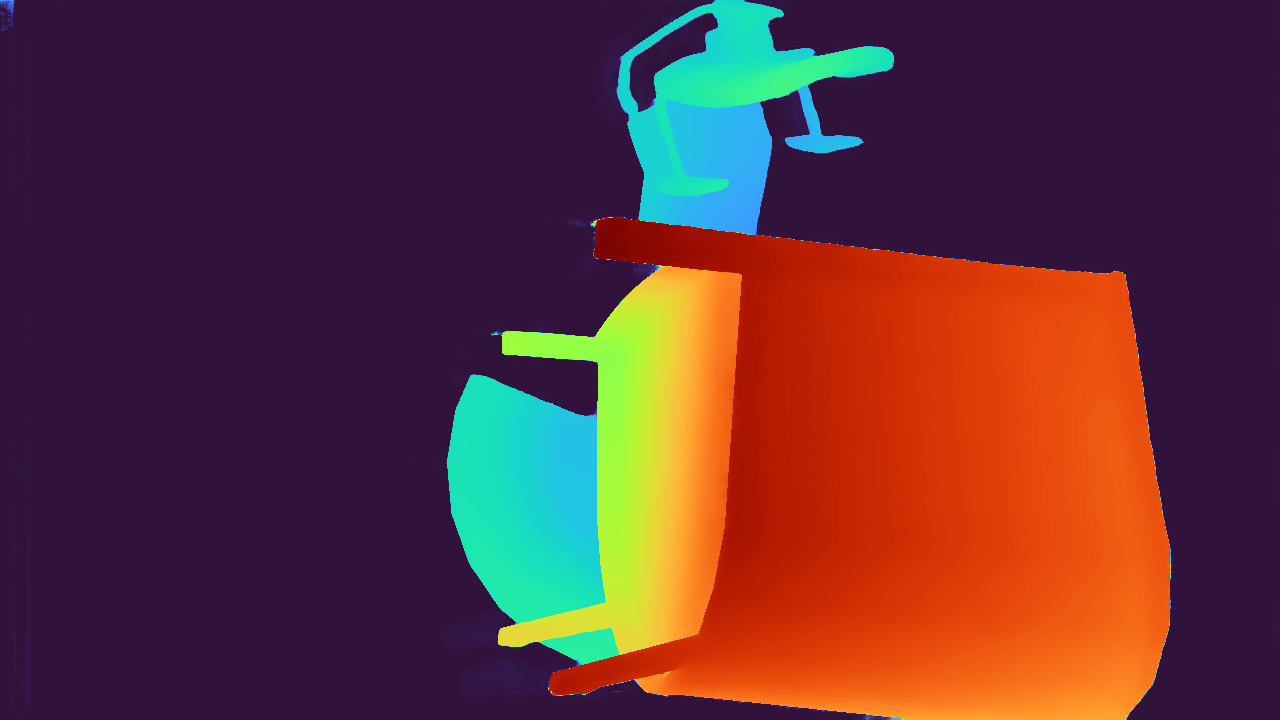}
    & \includegraphics[width=0.245\linewidth]{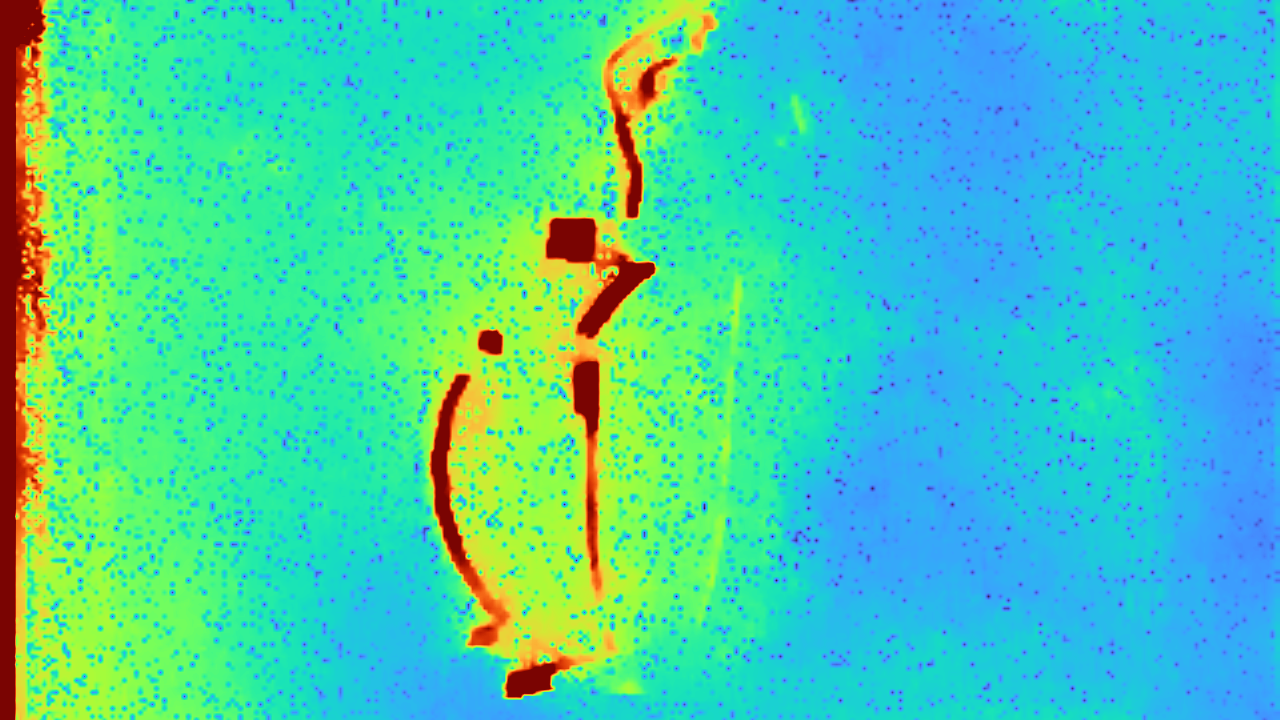}\\
     \includegraphics[width=0.245\linewidth]{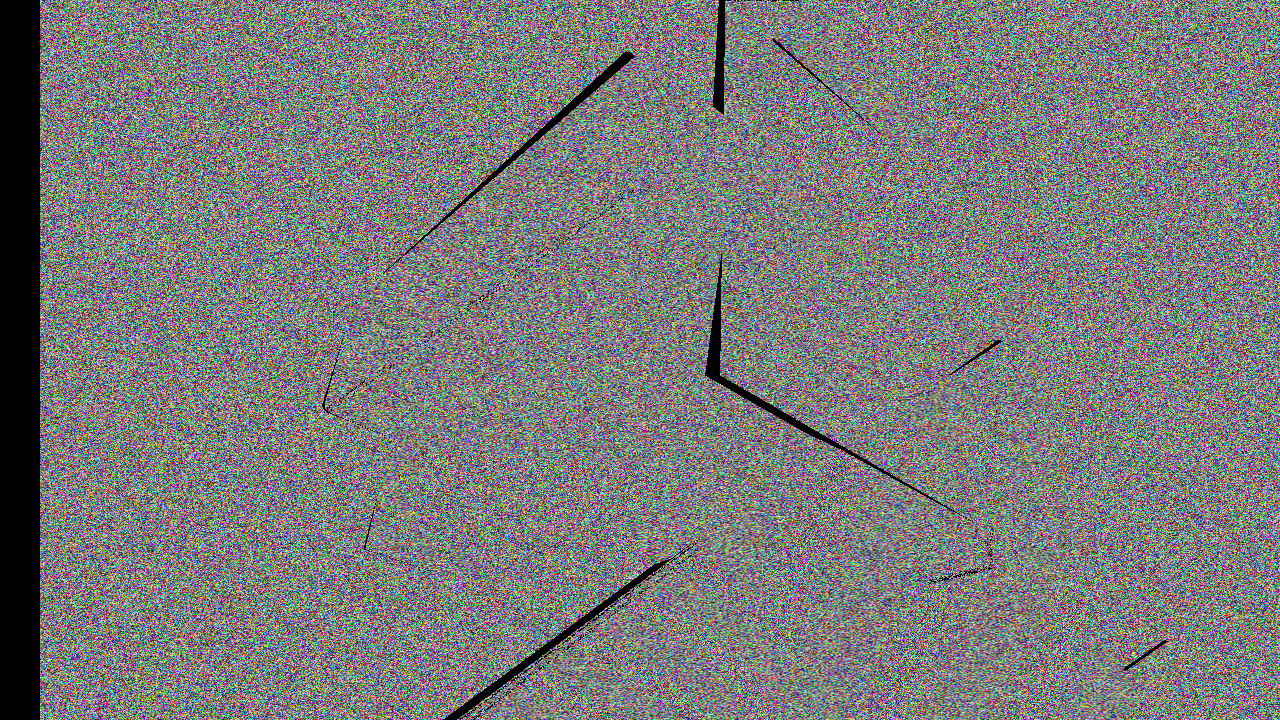}
    & \includegraphics[width=0.245\linewidth]{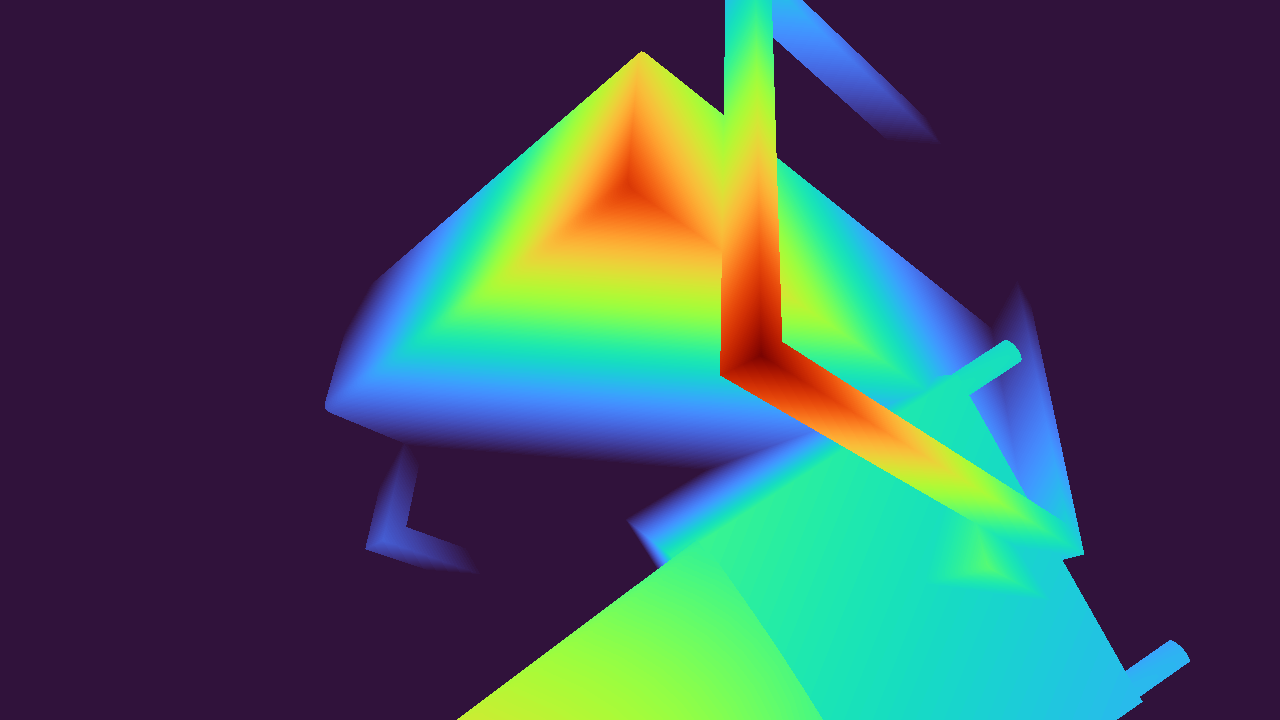}
    & \includegraphics[width=0.245\linewidth]{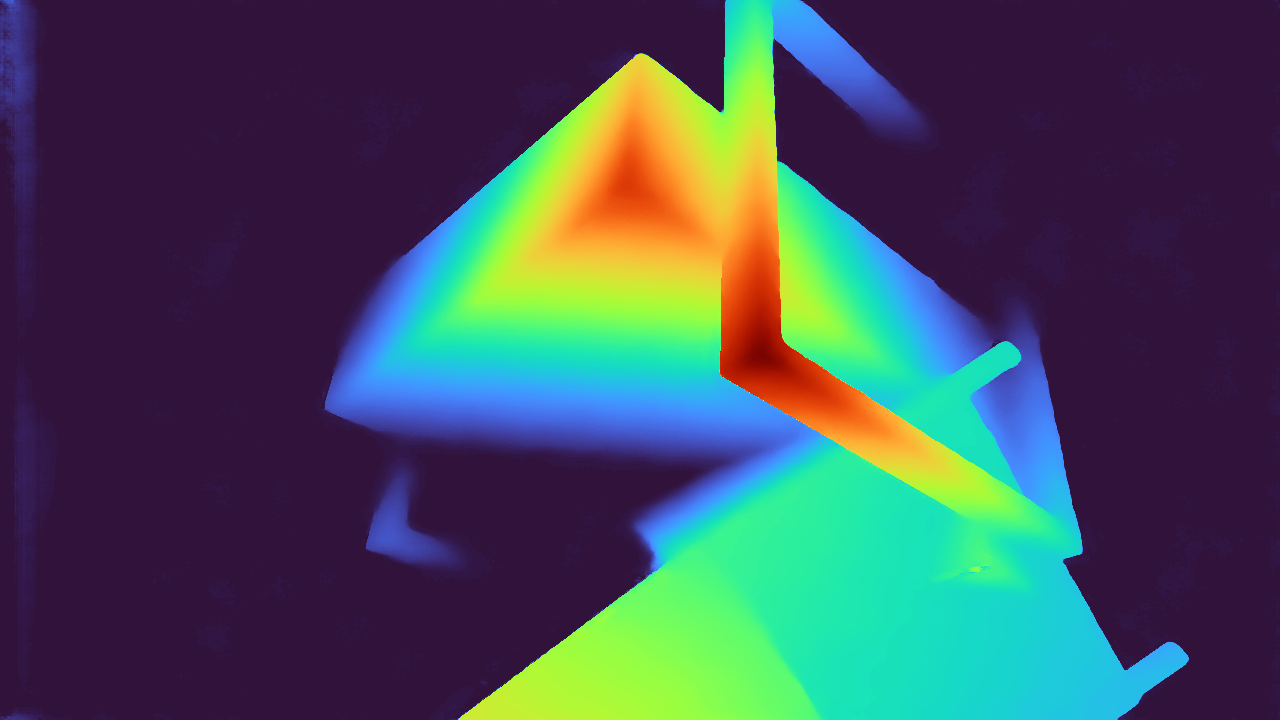}
    & \includegraphics[width=0.245\linewidth]{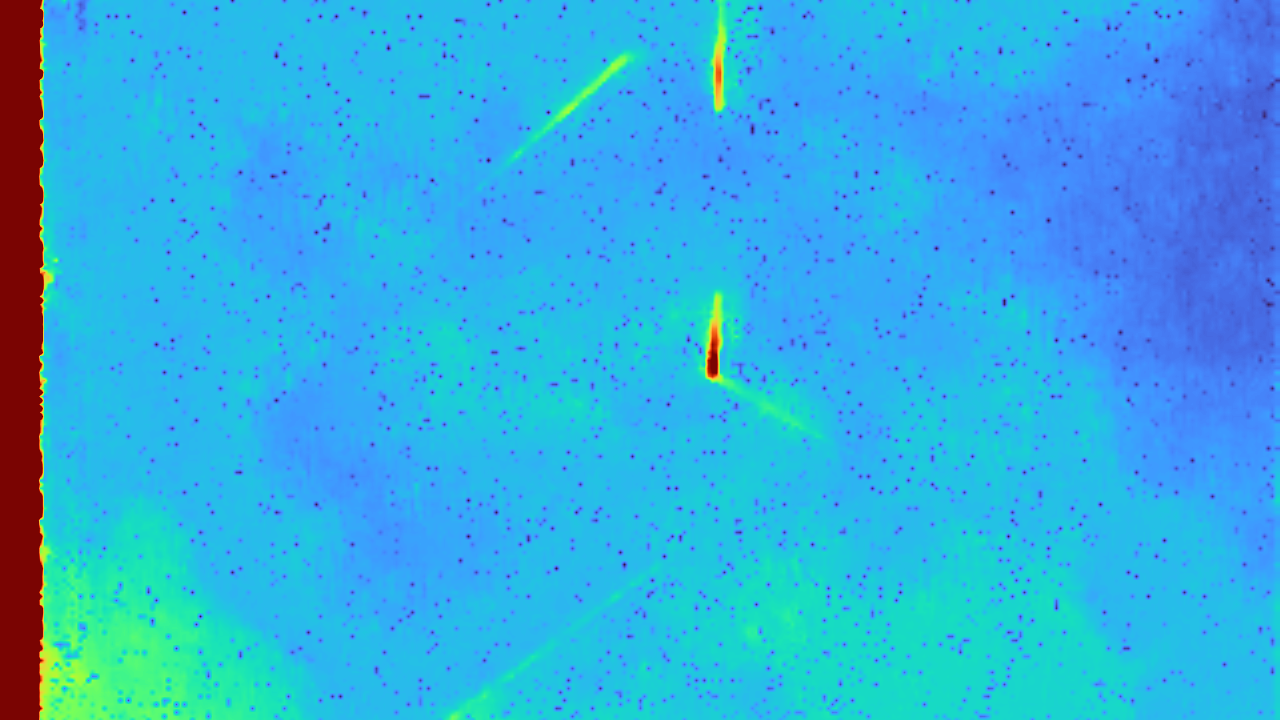}\\
    \end{tabular}
	\caption[Qualitative results on the RDS dataset.]{\textbf{Qualitative results on the Random Dot Stereo dataset.} Our network is able to recover disparities from RDS images while MADNet \cite{Tonioni_2019_CVPR} can not. The black holes in the left image is due to the occlusion between the left view and the cyclopean view \cite{Henkel1997}.}
    \label{fig:rd}
\end{figure}

To evaluate this claim, we created a Random Dot Stereogram (RDS) dataset. The dataset contains 2000 frames, 1800 for training and 200 for testing. We provide qualitative and quantitative results on the RDS dataset of Ours and MADNet \cite{Tonioni_2019_CVPR}. 
We fine-tuned Ours for 200 epochs with the ground truth. For MADNet, since it is a self-supervised method, we allow it to perform online fine-tuning for 200 epochs. 
The quantitative results are shown in Table~\ref{tab:RD}. 
Unlike MADNet, Ours successfully produces high-quality disparity maps, as shown in Figure~\ref{fig:rd}. 
To the best of our knowledge, our approach is the only end-to-end 2D convolution-based method that can pass the RDS test. This is because other 2D convolution-based methods are more rely on context information while our method is trying to find corresponding points with \emph{matching}. 

\begin{table}[t]
\centering
\scriptsize
\caption[Quantitative Results on RDS dataset.]{\textbf{Quantitative Results on RDS dataset}.}
\tabcolsep=0.35cm
\begin{tabular}{c c c c c}
\toprule
Methods & EPE & bad-1.0 & bad-2.0 & bad-3.0  \\ 
\hline
MADNet \cite{Tonioni_2019_CVPR} & 51.21  & 99.42  & 98.84 & 98.26\\ 
Ours  & 1.02   & 5.45  & 3.59 & 2.93 \\ 
\bottomrule
\end{tabular}
\label{tab:RD}
\end{table}

\section{Conclusion}
In this paper, we have proposed a highly efficient deep stereo matching network. Our method is not only on par with the state-of-the-art deep stereo matching methods in terms of accuracy, but is also able to run in super real-time, \ie, over 100 FPS on typical image sizes. This makes our method suitable for time-critical applications such as robotics and autonomous driving.  The key to our success is Displacement-Invariant Cost Computation, where 2D convolutions based cost computation is independently applied to all disparity levels to construct a 3D cost volume. In the future, we plan to extend this framework to the problem of optical flow estimation where a direct extension of volumetric methods will lead to a 5D feature volume and the requirement of 4D convolutions.

\noindent
\textbf{Acknowledgement}
We would like to thank Nikolai Smolyanskiy for helpful discussions. Y. Dai was supported in part by the Natural Science Foundation of China grants (61871325 and 61671387) and the National Key Research and Development Program of China (2018AAA0102803).

\vspace{-2mm}

\bibliographystyle{IEEEtran}
\bibliography{Stereo_Matching}

\end{document}